\documentclass[10pt,letterpaper,twocolumn]{article}
\usepackage{paper_style}
\usepackage{algorithm}
\usepackage{algorithmic}

\usepackage{caption}
\usepackage{graphicx}
\usepackage{microtype}
\usepackage{amsmath,amssymb,amsfonts,mathtools}
\usepackage{bm}
\usepackage{booktabs}
\usepackage{pifont}
\newcommand{\cmark}{\ding{51}}
\newcommand{\xmark}{\ding{55}}
\usepackage{booktabs}
\usepackage{multirow}
\anonymousfalse
\renewcommand{\paperauthors}{Ruiteng Zhao$^{12}$, Zhengshen Zhang$^{1}$, Yue Su$^{3}$, Wenshuo Wang$^{12}$, Jiahui Li$^{1}$, Zhiyuan Yang$^{2}$, \\Francis E.H. Tay$^{1}$, Marcelo H. Ang Jr.$^{1}$ and Haiyue Zhu$^{2\dagger}$}
\renewcommand{\paperaffiliation}{$^{1}$Advanced Robotics Centre, National University of Singapore, $^{2}$Singapore Institute of Manufacturing Technology, Agency for Science, Technology and Research (A*STAR), $^{3}$MMLab, The University of Hongkong}

\renewcommand{\paperprojectpage}{
  https://sg-wam.github.io/
}
\renewcommand{\papercorrespondence}{
  \href{mailto:your@email.com}{zhu\_haiyue@a-star.edu.sg}
}
\newcommand{\model}{SG-WAM}
\begin{document}

\maketopmatter
{SG-WAM: Self-Guided World Modeling in Geometry-Aware Policy Space}
{World Action Models (WAMs) couple action generation with prediction of future states. Their effectiveness depends on whether future dynamics are modeled in a space that is both aligned with action generation and sufficiently geometry-aware to capture where and how actions change the scene. Existing WAMs typically satisfy only part of this requirement, relying on either perceptually heavy observation-space targets or auxiliary latent spaces that are not jointly structured for action relevance and geometry. We propose \textbf{\model}, a self-guided framework that learns geometry-aware action-conditioned dynamics directly in the policy-derived representation space. {\model} introduces learnable dynamics tokens and a \textit{Self-Guided World Predictor} that forecasts their future latent states conditioned on intervening robot actions. Prediction targets are generated by an exponential moving average copy of the same policy backbone, providing stable supervision within the representation family used by the action expert. Geometric supervision further structures the policy image-token representations, providing spatially grounded context for the dynamics tokens and yielding a future-alignment space that is both action-relevant and geometry-aware. Latent future prediction, geometric grounding, and flow-matching action generation are jointly optimized end-to-end in a unified framework. Built on a 0.9B model without large-scale embodied pretraining, {\model} achieves 98.5\% average success on LIBERO and 73\% on LIBERO-Plus, while outperforming strong baselines in both in-distribution and out-of-distribution real-world evaluations.
SG-WAM}

\section{Introduction}

World Action Models (WAMs)~\cite{guo2026xwam,yuan2026fast,cen2025worldvla,su2026world,sun2026vla,bu2025univla,zhang2026dreamvla,chen2026dial} have recently emerged as a promising paradigm for robot manipulation by coupling action generation with prediction of future states. Unlike conventional policies that directly map observations to actions, WAMs explicitly supervise the policy to anticipate how robot interactions alter the environment and advance task progress. Such future-oriented supervision can encourage policy representations to capture action-induced dynamics that are weakly constrained by action imitation alone. However, the effectiveness of WAMs critically depends on the representation space in which future supervision is imposed and how closely this space is coupled to action generation. Existing WAMs~\cite{cen2025worldvla,cen2025rynnvla,zhang2026dreamvla,sun2026vla,yuan2026fast,hu2024vpp} broadly model action-conditioned futures in either observation space or latent representation space. Observation-space approaches predict future images, videos, depth maps, or other perceptual signals to provide dense supervision of scene evolution~\cite{cen2025worldvla,cen2025rynnvla,yang2026mantis}. Although informative, these targets require the model to capture variations in texture, viewpoint, background, etc., that may be only weakly related to the state changes governing successful manipulation. Consequently, substantial modeling capacity may be allocated to perceptual fidelity rather than manipulation-relevant dynamics.
\begin{figure}[!t]
    \centering
    \includegraphics[width=0.98\linewidth]{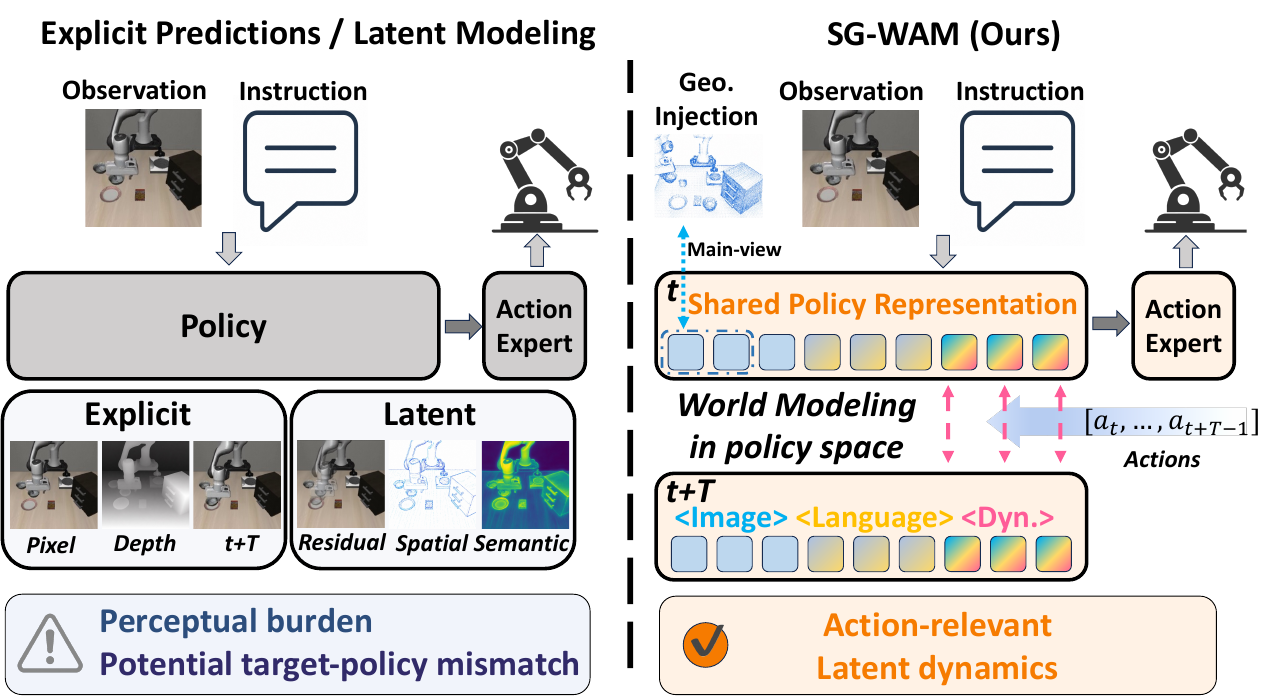}
    \caption{Conceptual comparison of existing WAMs and SG-WAM. Whereas explicit and auxiliary latent targets may introduce perceptual burden or target--policy mismatch, SG-WAM learns intervening-action-conditioned dynamics in geometry-structured, policy-derived representations.}
    \label{fig:intro_SGWAM}
\end{figure}

Latent WAMs instead model future information in representation space, reducing the burden of reconstructing manipulation-irrelevant appearance details. Existing methods instantiate this idea in different ways: some directly predict future visual features or align policy representations with features extracted from future observations~\cite{lyu2026lda1b,zheng2025flare,zhang2026dreamvla}, while others construct compact predictive representations designed to capture action-relevant feature changes or subspaces~\cite{sun2026vla,su2026world}. Despite their differences, these methods commonly impose future-oriented supervision through externally defined or auxiliary target representations that are not explicitly guaranteed to coincide with the representations directly used to condition action generation. This creates a potential mismatch between the information emphasized by the predictive objective and that required by the action expert.

Such limitations suggest that simply moving future prediction from observation space to latent space is insufficient. Although action-oriented representations can suppress appearance variation, such compression does not guarantee preservation of the fine-grained spatial information required for manipulation. An ideal future-modeling space must therefore not only be directly coupled to action generation, but also retain the geometry needed to represent where and how actions change the scene. This is critical because successful manipulation depends not only on recognizing what changes but also on capturing the spatial details~\cite{wang2025vggt,zhang2025falcon,peng2025omnivggt}. However, existing WAMs typically encode geometry through perceptual reconstruction, auxiliary targets, or predefined feature spaces~\cite{zhen2025tesseract,zhang2026dreamvla,guo2026xwam,sun2026vla,tian2026starry}, rather than directly using it to structure the policy-coupled future-alignment space. This raises a central challenge: \textit{how to construct a geometry-aware future-alignment space that is directly coupled to the representations used for action generation}.

\paragraph{Contribution} To address this challenge, we propose \textbf{\model}, a self-guided framework for learning geometry-aware action-conditioned dynamics directly in the policy representation space used by the action expert, as illustrated in Figure~\ref{fig:intro_SGWAM}. Rather than reconstructing future observations or predicting features from a separate future encoder, {\model} inserts learnable dynamics tokens into the VLM sequence and employs a \textit{Self-Guided World Predictor} to forecast their future states conditioned on intervening robot actions. The prediction targets are generated by an exponential moving average (EMA) copy of the same policy backbone, providing stable supervision within the same representation family used for action generation. To ensure that this policy-coupled future-alignment space retains the spatial structure required for manipulation, {\model} further uses geometric supervision to explicitly shape the policy image-token representations. These geometry-grounded image tokens provide spatial context for the dynamics tokens, yielding a latent dynamics space that is jointly action-relevant and geometry-aware. Latent future prediction, geometric grounding, and flow-matching action generation are optimized end-to-end in a unified training framework, allowing future supervision to directly organize the representations consumed by the action expert, while requiring only the current observation and language instruction at inference. 
\paragraph{Results} Built on a 0.9B model without large-scale embodied pretraining, {\model} achieves 98.5\% average success on LIBERO and 73\% on LIBERO-Plus, with clear gains over strong baselines in both in-distribution and out-of-distribution real-world evaluations.

\section{Related Work}

\paragraph{WAMs with Explicit Future Modeling}

Recent World Action Models extend Vision-Language-Action policies~\cite{bjorck2025gr00t,black2024pi_0,black2025pi05,kim2024openvla,zhang2025falcon,zhao2026fd,zhang2025falcon} by predicting how the environment evolves under robot actions~\cite{ye2026dreamzero,li2026lingbotva,hu2024vpp,cen2025worldvla}. A representative line of work generates future images or videos as intermediate targets for action generation. UniPi~\cite{du2023unipi} predicts task-conditioned videos and recovers actions through inverse dynamics, while subsequent methods improve long-horizon reasoning and controllability through vision-language planning, hierarchical decomposition, and stronger video generation models~\cite{du2024vlp,long2026vista,yang2025roboenvision,wan22,seedance2_2026}. Recent approaches further enrich predicted futures with depth, surface normals, or 4D structure ~\cite{zhen2025tesseract,guo2026xwam,zhou2026gem4d,wang2026mvista}, while structured representations such as optical flow and motion trajectories provide alternative interfaces between prediction and control ~\cite{ko2024avdc,xu2024Im2Flow2Act,zhi20253dflowaction,bharadhwaj2024gen2act}.
Although explicit prediction provides dense and interpretable supervision, observation-space objectives also require modeling texture, background, illumination, and viewpoint variations that may be weakly related to manipulation.

\begin{figure*}[t]
    \centering
    \includegraphics[width=\linewidth]{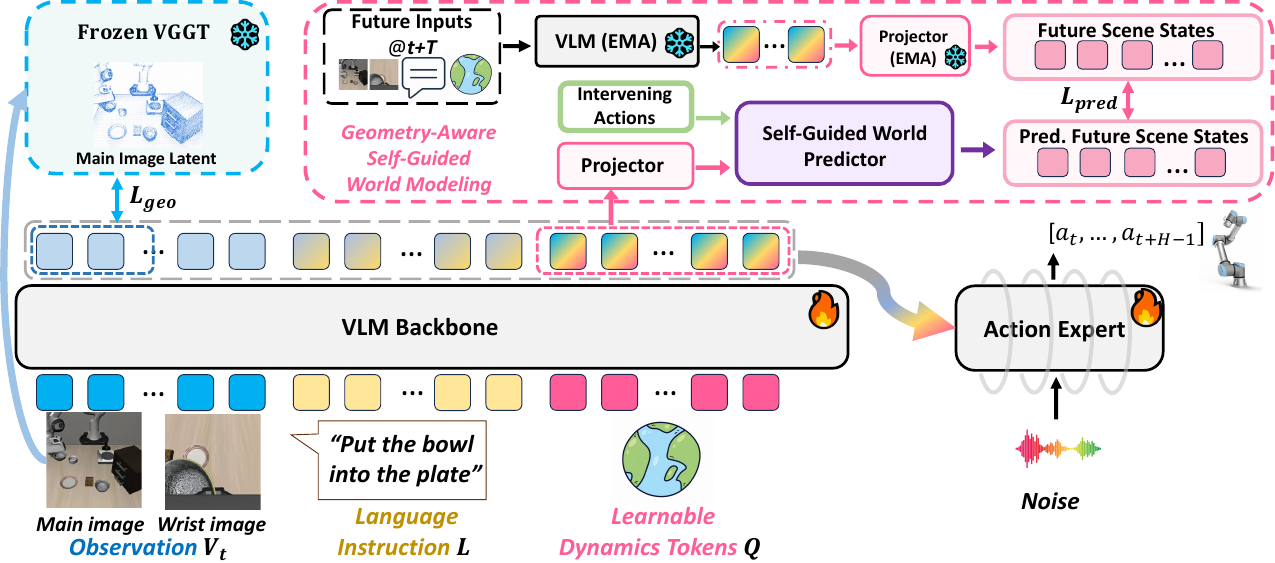}
    \caption{Overview of the {\model} framework. The VLM jointly contextualizes multi-view observations, language, and learnable dynamics tokens for action generation. During training, a frozen VGGT teacher shapes main-view image tokens, while SGWP predicts future dynamics-token states conditioned on intervening actions and aligns them with an EMA policy target. All auxiliary branches are removed at inference.}
    \label{fig:overview}
\end{figure*}

\paragraph{WAMs with Implicit Latent Modeling}
To avoid reconstructing detailed observations, another line of work models future dynamics in compact latent spaces. These methods learn predictive representations of scene evolution and integrate them with policy learning ~\cite{chen2025moto,bu2025univla,ye2025latent,chen2026villax,yuan2026fast}. Joint-embedding predictive models further show that future information can be learned without pixel-level reconstruction~\cite{assran2025v,maes2026leworldmodel}. In robot learning, latent future modeling has been instantiated through latent actions, predicted future features, imagined trajectories, and compact condition spaces ~\cite{bruce2024genie,gao2026dreamdojo,zhu2025irasim,zheng2025flare,su2026world,chen2026dial,sun2026vla}. Recent methods further explore future-informed latent queries, latent visual subgoals in pretrained feature spaces, and jointly modeled spatial-temporal latents and actions~\cite{luo2026being,chen2026lawam,tian2026starry}. Unified architectures increasingly couple predictive states with action generation through shared or closely connected policy pathways~\cite{bi2025motus,lyu2026lda1b}.

However, jointly training prediction and control does not necessarily mean that the future target is constructed from the same policy used for action generation. Existing WAMs obtain future supervision from future-observation or target encoders~\cite{zheng2025flare,sun2026vla}, pretrained feature spaces or latent visual subgoals~\cite{chen2026lawam}, future-informed posterior branches~\cite{luo2026being}, compact condition spaces~\cite{su2026world}, or jointly modeled spatial-temporal latents and actions~\cite{tian2026starry}. In contrast, SG-WAM constructs its current predictive state and future target from online and EMA copies of the same policy. Geometric supervision further shapes the policy visual context from which the dynamics-token states are formed, coupling spatial grounding, action-conditioned latent prediction, and action generation through shared policy representations.
\section{Method}

\subsection{Problem Formulation}

We seek to build a World Action Model (WAM) that jointly learns to generate robot actions and model how the environment evolves under those actions. Given the current multi-view observation $\bm{V}_t$ and task instruction $\bm{L}$, the model predicts an action chunk $\hat{\bm{A}}_t$ while learning how a latent policy state evolves under an intervening action sequence. During training, the expert action sequence
\begin{equation}
\bm{A}_t^{(\Delta)}
=
\left[
\bm{a}_t,\ldots,\bm{a}_{t+\Delta-1}
\right]
\end{equation}
conditions this transition, while the future observation
$\bm{V}_{t+\Delta}$ provides its target.

Let $\bm{z}_t^Q$ denote the latent dynamics representation associated with the current policy state, and let $\bar{\bm{z}}_{t+\Delta}^Q$ denote its future target representation. We model their action-conditioned transition as
\begin{equation}
\hat{\bm{z}}_{t+\Delta}^Q
=
\mathcal{F}_{\psi}
\left(
\bm{z}_t^Q,
\bm{e}_t^A
\right),
\qquad
\hat{\bm{z}}_{t+\Delta}^Q
\approx
\operatorname{sg}
\left(
\bar{\bm{z}}_{t+\Delta}^Q
\right),
\end{equation}
where $\bm{e}_t^A$ denotes the feature representation of  $\bm{A}_t^{(\Delta)}$, $\mathcal{F}_{\psi}$ denotes the action-conditioned transition model and $\operatorname{sg}(\cdot)$ denotes the stop-gradient operation. 

Unlike formulations that define the future target in observation space or through a separate future encoder, we construct both $\bm{z}_t^Q$ and $\bar{\bm{z}}_{t+\Delta}^Q$ from the same policy representation family. The following sections describe how these dynamics representations are constructed from policy token states, geometrically shaped, and used for both future prediction and action generation.

\subsection{Overall Framework}

{\model} realizes this formulation through two complementary designs. First, self-guided prediction models the action-conditioned evolution of policy-derived dynamics tokens using targets generated by the same policy pathway. Second, geometry supervision shapes the visual-token context from which these dynamics states are formed, encouraging the shared policy representation to preserve the spatial information required for manipulation.

As illustrated in Figure~\ref{fig:overview}, {\model} performs action-conditioned world modeling within the internal policy representation space. The visual, language, and learnable dynamics tokens are jointly contextualized within a shared VLM backbone (Qwen3.5-0.8B)~\cite{team2026qwen3}. Geometry supervision acts on the policy visual tokens, while the Self-Guided World Predictor models the action-conditioned evolution of the resulting dynamics-token states. These dynamics-token states remain in the context consumed by the action expert, allowing future prediction and action generation to jointly optimize the same policy representation.

Given the current multi-view observation $\bm{V}_t$ and language instruction
$\bm{L}$, they are first encoded into the corresponding representations $\bm{f}_t^{V}\in\mathbb{R}^{N_v\times D}$ and $\bm{f}^{L}\in\mathbb{R}^{N_l\times D}$, where $N_v$ and $N_l$ are the number of visual and language tokens, and $D$ is the feature dimension. We append $N_q$ learnable dynamics tokens $\bm{Q}\in\mathbb{R}^{N_q\times D}$ to the visual and language tokens and process them jointly through the VLM backbone $\mathcal{E}_{\theta}$ to get the last-layer contextualized token representations $\bm{H}_t$, which include visual, language and dynamics token states:
\begin{equation}
\bm{H}_t
=
\mathcal{E}_{\theta}
\left(
[\bm{f}_t^V,\bm{f}^L,\bm{Q}]
\right),
\qquad
\bm{H}_t
=
[\bm{H}_t^V,\bm{H}^L,\bm{H}_t^Q].
\end{equation}
Here, $\bm{H}_t^V$, $\bm{H}^L$, and $\bm{H}_t^Q$ denote the contextualized
visual, language, and dynamics-token states, respectively.

To construct a predictive space suitable for robust manipulation, a frozen 3D foundation model \cite{wang2025vggt} geometrically shapes the VLM main-view visual-token states $\bm{H}_t^{V,m}\in\mathbb{R}^{N_v^m\times D}$ during training, where $N_v^m$ denotes the number of main-view visual tokens. The Self-Guided World Predictor (SGWP) then takes the dynamics-token states $\bm H_t^Q \in \mathbb{R}^{N_q \times D}$, which are contextualized by the geometry-shaped visual states, together with the intervening action sequence $\bm{A}_t^{(\Delta)}$ to predict the future latent states. These predictions are aligned with the target EMA states~\cite{grill2020bootstrap} produced by feeding the future observation to an EMA copy of the policy.

Moreover, the contextualized VLM states $\bm{H}_t$ are also provided to a flow-matching action expert to generate the continuous action chunk $\hat{\bm{A}}_t$. During inference, the frozen geometry teacher, the SGWP prediction branch, and the EMA target pathway are removed. The online VLM, learnable dynamics tokens, and action expert form the deployed policy.

\subsection{Self-Guided World Modeling}

\paragraph{Geometry-Aware Policy States}
Although policy-derived latent representations are directly coupled to action generation, this coupling alone does not ensure that they preserve the local spatial information required for manipulation. We therefore use a frozen geometry model as an auxiliary teacher for the policy visual tokens. Importantly, the teacher features do not define the future-prediction target; they only shape the visual context from which the dynamics-token states are formed.

Given the current main-view observation $\bm{V}_t^m$, a frozen VGGT model extracts its final-layer geometric features, which are spatially pooled to obtain $\bm{Z}_t^{G}\in\mathbb{R}^{N_v^m\times D_g}$, where
$D_g$ denotes the VGGT feature dimension. On the policy side, a projector $\mathcal G_{\gamma}$ first maps the VLM last hidden states corresponding to the main-view visual tokens $\bm{H}_t^{V,m}$ to the teacher feature dimension $\hat{\bm{Z}}_t^{G}\in\mathbb{R}^{N_v^m\times D_g}$. We then align $\hat{\bm{Z}}_t^{G}$ with $\bm{Z}_t^{G}$ using the cosine similarity objective:
\begin{equation}
\mathcal{L}_{\mathrm{geo}}
=
\frac{1}{N_v^m}
\sum_{j=1}^{N_v^m}
\left[
1-
\cos
\left(
\hat{\bm{Z}}_{t,j}^{G},
\bm{Z}_{t,j}^{G}
\right)
\right].
\end{equation}

Because the dynamics tokens are jointly contextualized with these geometry-grounded visual states, $\mathcal{L}_{\mathrm{geo}}$ provides a pathway for spatial information to enter the policy-derived dynamics representation used for future prediction. See Appendix~\ref{app:geo} for more details.

\paragraph{Self-Guided World Predictor}
Built on the geometry-grounded policy context, the dynamics-token states serve as compact carriers for modeling action-conditioned scene evolution. We introduce the Self-Guided World Predictor (SGWP) to model how the scene state evolves under robot actions. The key distinction of the SGWP lies in its target construction. Instead of predicting future features defined by a separate encoder, both the current dynamics representation and its future target are derived from the same policy architecture. The online branch produces the current dynamics-token states, while an EMA copy of the policy produces a slowly evolving target from the future observation, allowing latent prediction to directly shape the policy representations used for action generation.

For the online pathway, the current dynamics-token states
$\bm{H}_t^{Q}\in\mathbb{R}^{N_q\times D}$ are mapped into the compact latent prediction space by the projector
${\bm{z}}_t^{Q}=\mathcal{P}_{\phi}(\bm{H}_t^{Q})$, $\bm{z}_t^{Q} \in \mathbb{R}^{N_q\times d}$, where $\phi$ denotes the parameters of the projector and $d$ is the space dimension. The intervening action sequence $\bm{A}_t^{(\Delta)}$ is encoded by an action encoder $\mathcal{A}_{\eta}$ as $\bm{e}_t^A=\mathcal{A}_{\eta}(\bm{A}_t^{(\Delta)})$.
Conditioned on the current dynamics representation and the intervening actions, SGWP predicts the future dynamics representation:
\begin{equation}
\hat{\bm{z}}_{t+\Delta}^Q
=
\mathcal{F}_{\psi}
\left(
\bm{z}_t^Q,
\bm{e}_t^A
\right),
\quad\hat{\bm{z}}_{t+\Delta}^{Q}\in\mathbb{R}^{N_q\times d},
\end{equation}
where $\mathcal{F}_{\psi}$ denotes the action-conditioned predictor. See Appendix~\ref{app:sgwp} for more details of SGWP.

The prediction target is generated from the future observation
$\bm{V}_{t+\Delta}$ and the same language instruction $\bm{L}$ by an EMA copy of the same policy pathway. Specifically, the target VLM processes the future visual tokens $\bm{f}_{t+\Delta}^{V}\in\mathbb{R}^{N_v\times D}$, language tokens $\bm{f}^{L}\in\mathbb{R}^{N_l\times D}$, and EMA dynamics tokens $\bar{\bm{Q}}\in\mathbb{R}^{N_q\times D}$ to obtain
\begin{equation}
\bar{\bm{H}}_{t+\Delta}^Q
=
\left[
\mathcal{E}_{\bar{\theta}}
\left(
[
\bm{f}_{t+\Delta}^{V},
\bm{f}^{L},
\bar{\bm{Q}}
]
\right)
\right]^Q,
\end{equation}
where $[\cdot]^Q$ selects the states corresponding to the dynamics tokens. The target states are then projected into the same latent prediction space: $\bar{\bm{z}}_{t+\Delta}^Q=\mathcal{P}_{\bar{\phi}}(\bar{\bm{H}}_{t+\Delta}^Q)$, $\bar{\bm{z}}_{t+\Delta}^{Q}\in\mathbb{R}^{N_q\times d}$.
The target parameters are updated by
\begin{equation}
\bar{\Theta}
\leftarrow
\mu \bar{\Theta}
+
(1-\mu)\Theta,
\end{equation}
where $\Theta=(\theta,\phi,Q)$ and $\bar{\Theta}=(\bar{\theta},\bar{\phi},\bar{Q})$ denote the online and EMA parameters of the VLM $\mathcal{E}$, the projector $\mathcal{P}$, and the dynamics tokens, and $\mu$ is the EMA momentum coefficient. The target pathway is detached
from gradient backpropagation and updated only through the EMA rule.

The latent prediction objective minimizes the mean squared error (MSE) between the predicted and EMA future states:
\begin{equation}
\mathcal{L}_{\mathrm{pred}}
=
\frac{1}{N_q d}
\left\|
\hat{\bm{z}}_{t+\Delta}^Q
-
\operatorname{sg}
\left(
\bar{\bm{z}}_{t+\Delta}^Q
\right)
\right\|_F^2,
\end{equation}
where $\operatorname{sg}(\cdot)$ denotes stop-gradient. Gradients from $\mathcal{L}_{\mathrm{pred}}$ update the online VLM, dynamics
tokens, projector, action encoder, and SGWP, while leaving the EMA target
pathway gradient-free. Because the same dynamics-token states remain in the context supplied to the action expert, the prediction and action objectives jointly optimize the shared policy backbone.

\begin{table*}[t]
\centering
\small
\setlength{\tabcolsep}{2.5pt}
\begin{tabular*}{\linewidth}{@{\extracolsep{\fill}}lccccccc}
\toprule
Method & Params & Embodied PT. & Spatial & Object & Goal & Long & Avg. \\
\midrule
OpenVLA-OFT~\cite{kim2025fine} & 7B & \cmark & 97.6 & 98.4 & 97.9 & 94.5 & 97.1 \\
$\pi_0$~\cite{black2024pi_0} & 3.3B & \cmark & 98.0 & 96.8 & 94.4 & 88.4 & 94.4 \\
$\pi_0$-FAST~\cite{pertsch2025fast} & 3.3B & \cmark & 96.4 & 96.8 & 88.6 & 60.2 & 85.5 \\
$\pi_{0.5}$~\cite{black2025pi05} & 3.3B & \cmark & 98.8 & 98.2 & 98.0 & 92.4 & 96.9 \\
GR00T N1.6~\cite{bjorck2025gr00t} & 3B & \cmark & 97.7 & 98.5 & 97.5 & 94.4 & 97.0 \\
Spatial Forcing~\cite{li2025spatial} & 7B & \cmark & \textbf{99.4} & 99.6 & \textbf{98.8} & \underline{96.0} & \textbf{98.5} \\
WorldVLA~\cite{cen2025worldvla} & 7B & \xmark & 87.6 & 96.2 & 83.4 & 60.0 & 81.8 \\
LAPA~\cite{ye2025latent} & 7B & \cmark & 55.4 & 58.8 & 74.6 & 73.8 & 65.7 \\
RynnVLA-002~\cite{cen2025rynnvla} & 7B & \xmark & \underline{99.0} & \underline{99.8} & 96.4 & 94.4 & 97.4 \\
Mantis~\cite{yang2026mantis} & 5.8B & \cmark & 98.8 & 99.2 & 94.4 & 94.2 & 96.7 \\
UniVLA~\cite{bu2025univla} & 7B & \cmark & 96.5 & 96.8 & 95.6 & 92.0 & 95.2 \\
Fast-WAM~\cite{yuan2026fast} & 6B & \xmark & 98.2 & \textbf{100.0} & 97.0 & 95.2 & \underline{97.6} \\
VLA-JEPA~\cite{sun2026vla} & 2B & \xmark & 94.8 & 99.6 & 95.8 & 94.0 & 96.1 \\
\midrule
\textbf{\model} & \textbf{0.9B} & \xmark & \textbf{99.4} & \underline{99.8} & \underline{98.6} & \textbf{96.2} & \textbf{98.5} \\
\bottomrule
\end{tabular*}
\caption{Simulation results on LIBERO. Success rates are reported for the four standard LIBERO suites. \textbf{Bold} indicates the best result, and \underline{underlining} indicates the second-best result.}
\label{tab:libero_results}
\end{table*}

\begin{table*}[t]
\centering
\small
\setlength{\tabcolsep}{2.5pt}
\begin{tabular*}{\linewidth}{@{\extracolsep{\fill}}lccccccccc}
\toprule
Method & Params & Camera & Robot & Language & Light & Background & Noise & Layout & Overall \\
\midrule
WorldVLA~\cite{cen2025worldvla} & 7B & 0.1 & 27.9 & 41.6 & 43.7 & 17.1 & 10.9 & 38.0 & 25.0 \\
Spatial Forcing~\cite{li2025spatial} & 7B & 20.1 & 13.4 & 40.9 & 29.1 & 33.4 & 25.7 & 39.3 & 29.1 \\
Mantis~\cite{yang2026mantis} & 5.8B & 15.7 & 41.8 & 45.9 & 45.1 & 28.9 & 39.2 & 62.5 & 39.8 \\
UniVLA~\cite{bu2025univla} & 7B & 4.3 & \underline{50.3} & 71.8 & 59.1 & 80.0 & 25.3 & 34.3 & 41.5 \\
Fast-WAM~\cite{yuan2026fast} & 6B & 16.4 & 44.5 & 68.9 & 78.2 & 53.7 & 37.7 & 60.7 & 50.0 \\
$\pi_0$~\cite{black2024pi_0} & 3.3B & 13.8 & 6.0 & 58.8 & 85.0 & 81.4 & \underline{79.0} & 68.9 & 53.6 \\
VLA-JEPA~\cite{sun2026vla} & 2B & 40.3 & \textbf{55.7} & 72.9 & 88.2 & 70.5 & 38.2 & \textbf{74.6} & 62.9 \\
OpenVLA-OFT~\cite{kim2025fine} & 7B & \underline{56.4} & 31.9 & \underline{79.5} & \underline{88.7} & \textbf{93.3} & 75.8 & \underline{74.2} & \underline{69.6} \\
\midrule
\textbf{\model} & \textbf{0.9B} & \textbf{58.6} & 48.9 & \textbf{81.4} & \textbf{89.8} & \underline{86.1} & \textbf{80.7} & \underline{74.2} & \textbf{73.0} \\
\bottomrule
\end{tabular*}
\caption{Zero-shot transfer results on LIBERO-Plus. Success rates are reported under different perturbation settings. \textbf{Bold} indicates the best result, and \underline{underlining} indicates the second-best result.}
\label{tab:libero_plus_results}
\end{table*}

\subsection{Conditional Flow-Matching Action Expert}

The action expert is conditioned on the complete policy context $\bm{H}_t=[\bm{H}_t^V,\bm{H}^L,\bm{H}_t^Q]$, such that the dynamics-token states optimized by future prediction remain part of the context used for action generation. We adopt conditional flow matching to generate an $H$-step action chunk. Given an expert action chunk $\bm{A}_t$, Gaussian noise $\bm{\epsilon}\sim\mathcal{N}(\bm{0},\bm{I})$, and $\tau\sim\mathcal{U}(0,1)$, we define $\bm{A}_t^\tau=\tau\bm{A}_t+(1-\tau)\bm{\epsilon}$ and optimize
\begin{equation}
\mathcal{L}_{\mathrm{act}}
=
\mathbb{E}_{\bm{A}_t,\bm{\epsilon},\tau}
\left[
\left\|
v_{\omega}
\left(
\bm{A}_t^{\tau},
\tau,
\bm{H}_t
\right)
-
\left(
\bm{A}_t-\bm{\epsilon}
\right)
\right\|_2^2
\right].
\end{equation}
At inference, the action chunk is obtained by integrating the learned velocity field while conditioning on $\bm{H}_t$.

\subsection{Training Objective}

The model is jointly trained with
\begin{equation}
\mathcal{L}
=
\mathcal{L}_{\mathrm{act}}
+
\lambda_{\mathrm{geo}}\mathcal{L}_{\mathrm{geo}}
+
\lambda_{\mathrm{pred}}\mathcal{L}_{\mathrm{pred}},
\end{equation}
where $\lambda_{geo}$ and $\lambda_{pred}$ weight the geometry-shaping and action-conditioned latent prediction losses.
The action loss updates the action expert and the shared online policy backbone, including the dynamics tokens. The prediction loss additionally updates the prediction projector, action encoder, and SGWP, while the geometry loss updates the geometry projector and shapes the online visual-token states. The VGGT teacher remains frozen, and the target policy pathway is updated only through EMA. During inference, the geometry teacher, SGWP prediction branch, and EMA target pathway are removed. All trainable online components are jointly optimized in a single training stage.

\section{Experiments}

To evaluate the generalization and robustness of SG-WAM, we conduct comprehensive simulation and real-world experiments. We use two simulation benchmarks, LIBERO and LIBERO-Plus, and further evaluate the policy on real-world robot manipulation tasks. We compare SG-WAM with representative baselines to assess its performance across both simulated and physical environments. We provide more details of simulation and real-world experiments in Appendix~\ref{app:sim} and Appendix~\ref{app:real}.


\subsection{Simulation Setup and Baselines}

LIBERO is a standard benchmark for robot manipulation. It contains four task suites, LIBERO-Spatial, LIBERO-Object, LIBERO-Goal, and LIBERO-Long, which evaluate spatial reasoning, object-centric manipulation, goal-conditioned behavior, and long-horizon task execution. LIBERO-Plus extends this setting with more challenging task configurations and visual variations, providing a stronger test of policy robustness and generalization.
For LIBERO, we jointly train the policy on all four standard suites and evaluate it on the corresponding benchmark tasks. For LIBERO-Plus, we use the policy trained on LIBERO and directly evaluate it without fine-tuning, forming a zero-shot transfer setting. This protocol allows us to assess both in-domain manipulation performance and out-of-distribution generalization under more challenging task variations.

We compare {\model} with representative baselines, including generalist pretrained policies~\cite{kim2025fine,black2024pi_0,black2025pi05,pertsch2025fast,bjorck2025gr00t,li2025spatial}, explicit WAMs~\cite{cen2025worldvla,cen2025rynnvla,yang2026mantis} and implicit WAMs~\cite{ye2025latent, bu2025univla,yuan2026fast,sun2026vla}. These comparisons cover the main families of recent approaches. Simulation baseline results are reported from the corresponding papers under their stated LIBERO and LIBERO-Plus protocols.


\subsection{Simulation Results}

Table~\ref{tab:libero_results} reports the results on the four standard LIBERO suites. SG-WAM achieved an average success rate of 98.5\%, matching the strongest baseline while using a substantially smaller 0.9B model and no additional embodied pretraining. In contrast, most high-performing baselines rely on larger backbones or large-scale embodied pretraining. 
LIBERO-Plus evaluates zero-shot robustness under distribution shifts in camera viewpoint, robot embodiment, language, illumination, background, observation noise, and scene layout. As shown in Table~\ref{tab:libero_plus_results}, SG-WAM achieves the highest overall success rate of 73.0\%, demonstrating the strongest overall zero-shot transfer performance across these perturbations. Although Spatial Forcing and Fast-WAM are among the strongest methods on standard LIBERO, their performance decreases substantially under the distribution shifts introduced by LIBERO-Plus. In contrast, SG-WAM achieves the best performance under camera, language, illumination, and layout shifts while remaining competitive under background and observation-noise perturbations, suggesting that its learned policy representations are less sensitive to changes in visual and linguistic input distributions. These results indicate that SG-WAM provides a stronger balance between in-distribution task performance and out-of-distribution robustness.

\subsection{Real-World Experimental Setup and Baselines}

\begin{figure}[t]
    \centering
    \includegraphics[width=\linewidth]{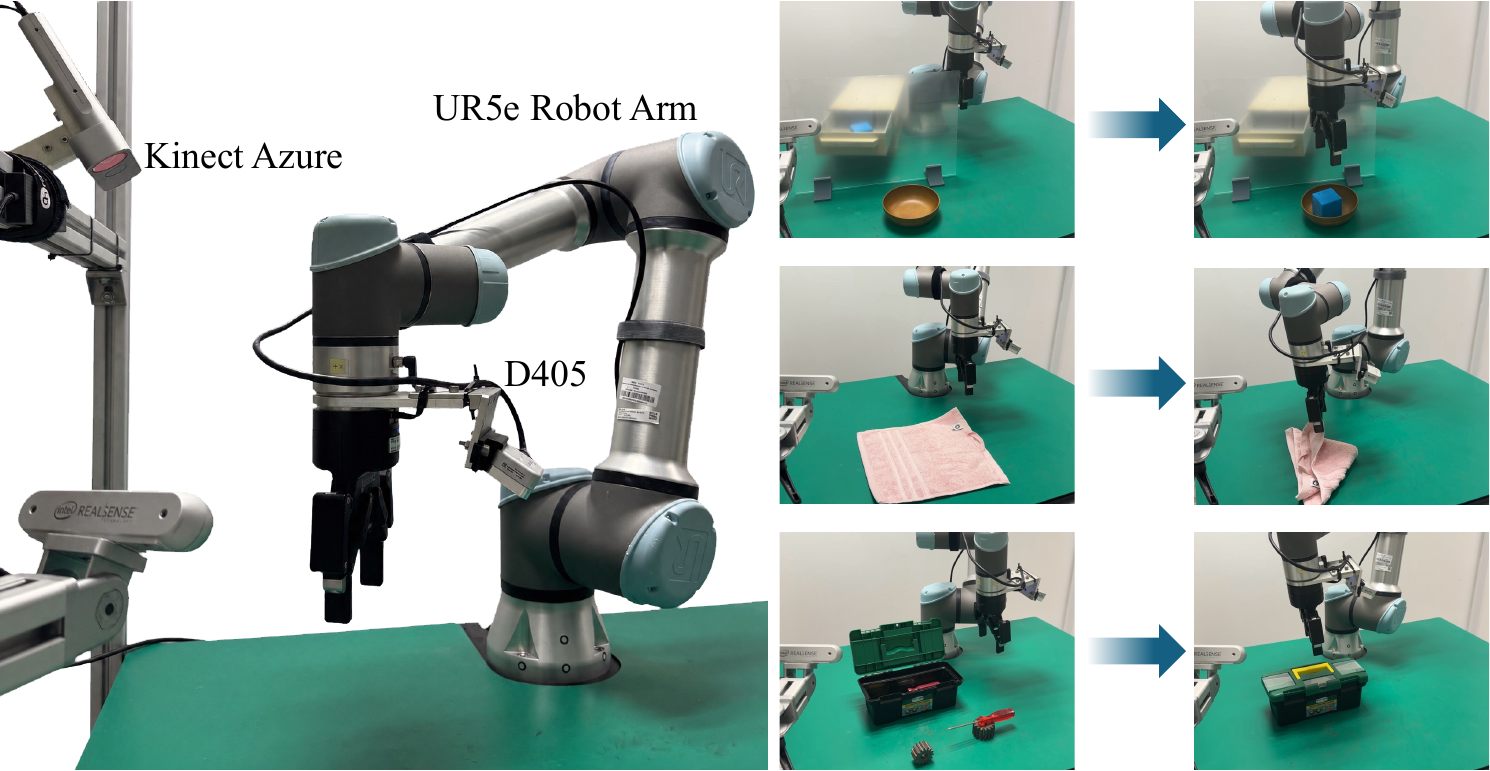}
    \caption{\textbf{Left:} Visualization of the real-world platform. We use a UR5e robot arm as the manipulation platform, the Kinect Azure camera as the main camera and the RealSense D405 as the gripper camera. \textbf{Right:} Visualization of 3 real-world tasks. 1) \textbf{Top:} Pick and Place. 2) \textbf{Middle:} Towel Folding. 3) \textbf{Bottom:} Toolbox Organization.}
    \label{fig:realworld_setup}
\end{figure}

We construct the in-distribution (ID) setting using three tasks and collect 100 expert demonstrations for each task:
(1) \textbf{Pick and Place.} The robot retrieves a blue cube from an open drawer and places it into a bowl while a barrier obstructs the direct transfer path. 
(2) \textbf{Towel Folding.} The robot folds a flat towel in half twice to produce a compact configuration. 
(3) \textbf{Toolbox Organization.} The robot sequentially places a screwdriver and two gears into a toolbox before closing it. 
Figure~\ref{fig:realworld_setup} shows the visualization of the tasks.
To evaluate generalization, we assess out-of-distribution (OOD) generalization under three conditions not presented in the expert data: \textbf{Background Shift}, \textbf{Light Change}, and \textbf{Novel Object}.



We use a UR5e robotic arm as our main manipulation platform, an Azure Kinect camera to capture main RGB image, and a RealSense D405 to capture robot gripper RGB image. The real experimental platform is shown in Figure~\ref{fig:realworld_setup}. 
We select VPP~\cite{hu2024vpp} and VLA-JEPA~\cite{sun2026vla} as representative baselines for our real-world experiments, covering explicit and implicit world-modeling paradigms, respectively. VPP explicitly models future visual observations through video prediction, whereas VLA-JEPA predicts future representations in latent space without reconstructing observations.

For the real-world evaluation, we test each method for 20 trials on Pick and Place and Towel Folding tasks. Because the Toolbox Organization task requires substantially longer execution, we conduct 10 trials per method on this task. For each trial, the initial scene is randomized and reproduced as closely as possible across methods to ensure a fair comparison.

\begin{table*}[t]
\centering
\small
\setlength{\tabcolsep}{2.5pt}
\begin{tabular}{@{}l cccc cccc c@{}}
\toprule
\multicolumn{1}{c}{\multirow[c]{2}{*}{\textbf{Model}}}
& \multicolumn{4}{c}{\textbf{Pick and Place}}
& \multicolumn{4}{c}{\textbf{Towel Folding}}
& \multicolumn{1}{c}{\textbf{Toolbox Organization}} \\
\cmidrule(lr){2-5}
\cmidrule(lr){6-9}
\cmidrule(l){10-10}
& ID & Background & Light Change & Novel Object & ID & Background & Light Change & Novel Object & ID \\
\midrule
VLA-JEPA & 35\% & 20\% & 25\% & 20\% & 20\% & 10\% & 10\% & 15\% & 20\% \\
VPP & 30\% & 15\% & 10\% & 10\% & 35\% & 15\% & 15\% & 10\% & 30\% \\
\midrule
\textbf{\model} & \textbf{75\%} & \textbf{55\%} & \textbf{60\%} & \textbf{40\%} & \textbf{45\%} & \textbf{25\%} & \textbf{35\%} & \textbf{25\%} & \textbf{50\%} \\
\bottomrule
\end{tabular}
\caption{Success rates under different visual perturbations. \textbf{Bold} indicates the best result.}
\label{tab:realworld_id_ood}
\end{table*}

\subsection{Real-World Results}

 As shown in Table~\ref{tab:realworld_id_ood}, {\model} consistently outperformed VPP and VLA-JEPA under both in-distribution (ID) and out-of-distribution (OOD) conditions. It achieved the highest ID success rate across all three tasks and retained its advantage under changes in background, light, and object. This advantage also extended to the long-horizon Toolbox Organization task, whose multi-stage structure increases the risk of compounding execution errors. These results indicate that {\model} supports reliable execution under familiar conditions while remaining robust to visual distribution shifts and extended manipulation horizons.

\subsection{Ablation Study}

We ablate geometric supervision and self-guided world modeling on the four LIBERO suites. As shown in Table~\ref{tab:ablation_module}, the model without either component achieves an average success rate of 95.3\%. 
Combining both components yields the best average performance of 98.5\%. 
Removing world modeling reduces the average success rate by 1.9 percentage points, compared with a 0.9-percentage-point reduction after removing geometric supervision. This difference is most pronounced on LIBERO-Long, where removing world modeling decreases performance from 96.2\% to 92.2\%. 
These results provide empirical support for the effectiveness of self-guided world modeling in action generation, especially on tasks that require coherent modeling of long-horizon state transitions. Geometric supervision grounds the learned representations in manipulation-relevant spatial structure, thereby facilitating the organization of action-conditioned transitions. Together, the two objectives produce the strongest performance, particularly on LIBERO-Long, suggesting that spatial grounding and predictive supervision provide complementary benefits.

\begin{table}[t]
\centering
\small
\begin{tabular*}{\columnwidth}
{@{\extracolsep{\fill}}cc|ccccc@{}}
\toprule
Geo. & WM. & Spatial & Object & Goal & Long & Avg. \\
\midrule
\xmark & \xmark
& 97.0 & 97.6 & 95.6 & 91.0 & 95.3 \\

\cmark & \xmark
& 98.2 & 97.8 & 98.0 & 92.2 & 96.6 \\

\xmark & \cmark
& 97.8 & \textbf{99.8} & 98.2 & 94.4 & 97.6 \\

\cmark & \cmark
& \textbf{99.4} & \textbf{99.8} & \textbf{98.6}
& \textbf{96.2} & \textbf{98.5} \\
\bottomrule
\end{tabular*}
\caption{Ablation study of geometric supervision and
self-guided world modeling on LIBERO. Geo. denotes geometric
supervision, and WM. denotes self-guided world modeling.
Success rates (\%) are reported. \textbf{Bold} indicates the
best result.}
\label{tab:ablation_module}
\end{table}

\begin{table}[t]
\centering
\small
\begin{tabular*}{\columnwidth}
{@{\extracolsep{\fill}}c|ccccc@{}}
\toprule
Number of Token & Spatial & Object & Goal & Long & Avg. \\
\midrule
1  & 97.2 & 98.8 & 98.2 & 90.2 & 96.1 \\
4  & 99.0 & 98.4 & 97.4 & 95.2 & 97.5 \\
8  & \textbf{99.4} & \textbf{99.8} & \textbf{98.6}
   & \textbf{96.2} & \textbf{98.5} \\
16 & \textbf{99.4} & 99.4 & 97.6 & 92.4 & 97.2 \\
\bottomrule
\end{tabular*}
\caption{Ablation study on the number of learnable dynamics
tokens on LIBERO. Success rates (\%) are reported.
\textbf{Bold} indicates the best result.}
\label{tab:ablation_num_token}
\end{table}

We further varied the number of learnable dynamics tokens while keeping all other settings fixed. Increasing the number of tokens from one to eight improved the average success rate from 96.1\% to 98.5\%. The largest improvement occurred on LIBERO-Long, where the success rate increased from 90.2\% to 96.2\%. However, increasing the token number to 16 reduced the average success rate to 97.2\%. We therefore used eight dynamics tokens in the final model. This non-monotonic trend suggests that the gain is not simply proportional to the number of dynamics tokens. Instead, a moderate number of tokens appears sufficient to represent diverse interaction dynamics, whereas additional tokens provide no further benefit to action learning.

\begin{figure}[!t]
    \centering
    \includegraphics[width=\linewidth]{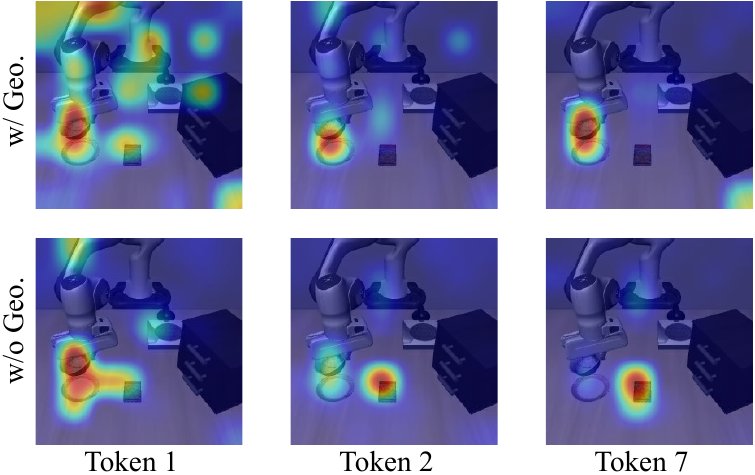}
    \caption{Representative middle-layer attention maps from dynamics tokens to main-view image tokens. The task is \textit{Pick up the black bowl between the plate and the ramekin and place it on the plate.} With geometric supervision, the selected tokens attend more consistently to the robot–object interaction regions.}
    \label{fig:attention_map}
\end{figure}

To further explore how geometric supervision influences the visual context used for latent dynamics modeling, we visualize the middle-layer attention from each dynamics token to the main-view image tokens from the same frame in Figure~\ref{fig:attention_map}. With geometric supervision, the dynamics tokens attend more consistently to interaction-relevant regions, including the robot end effector, the manipulated receptacle, and their surrounding spatial context. Without geometric supervision, the attention is more frequently dominated by isolated visually salient objects, such as the package on the table, and shows weaker correspondence with the robot-object interaction. These patterns suggest that geometric supervision helps organize the policy-derived representation space around manipulation-relevant spatial relationships, providing a more structured visual context for action-conditioned self-guided world modeling. Complete attention maps and the null-action ablation are provided in Appendix~\ref{app:ablation}.

\section{Conclusion}
This paper introduced {\model}, a self-guided world action model that learns action-conditioned future dynamics in a compact latent prediction space derived from the policy itself. {\model} introduces learnable dynamics tokens whose future states are predicted from the current policy state and intervening robot actions, while an EMA copy of the same policy provides stable future targets. A frozen 3D foundation model further shapes the policy's main-view visual tokens with geometric structure, allowing latent future prediction and action generation to be jointly optimized over spatially grounded policy representations. The geometry teacher, SGWP, and EMA target pathway are used only during training and are removed at inference. Without large-scale embodied pretraining, the 0.9B-parameter {\model} achieves 98.5\% average success on LIBERO and 73.0\% on LIBERO-Plus, while outperforming representative explicit and latent WAM baselines in real-world evaluations. These results indicate that coupling self-guided future prediction with geometric grounding provides an effective approach to learning robust policies. Future work will investigate scaling this self-guided world-modeling framework to larger policy backbones and broader cross-embodiment datasets, with the goal of supporting diverse embodiments and manipulation tasks.

\clearpage
\bibliographystyle{plainnat}
\bibliography{references}

\FloatBarrier
\appendix
\onecolumn
\section*{Appendix}
\large
\section{Details of Learnable Dynamics Tokens}
We introduce the learnable dynamics tokens $\bm{Q}\in\mathbb{R}^{N_q\times D}$ to represent the action-relevant latent state of the observation. $D$ is the hidden dimension of the vision-language backbone and $N_q=8$ is the token number. The dynamics tokens $Q$ are directly optimized continuous embeddings. After the backbone constructs the text embeddings and merges the visual embeddings into the image placeholder positions, we append the dynamics token embeddings to the resulting sequence $[\bm{f}_t^{V},\bm{f}^{L},\bm{Q}]$,
where $\bm{f}_t^{V}\in\mathbb{R}^{N_v\times D}$ and $\bm{f}^{L}\in\mathbb{R}^{N_l\times D}$ denote the corresponding visual and language embeddings. The attention mask is extended with $N_{q}$ valid positions. The multimodal rotary position indices of the dynamics tokens $\bm{Q}$ continue from the last position of the original image-language sequence. $\bm{Q}$ therefore participate in every transformer layer and aggregate information from the visual observations and language instruction.

The final-layer VLM states $\bm{H}_t$ is passed to the action expert: $\bm{H}_t=\mathcal{E}_{\theta}\left([\bm{f}_t^{V},\bm{f}^{L},\bm{Q}]
\right)$, where $\mathcal{E}_{\theta}$ is the VLM backbone. Thus, the learnable dynamics tokens $\bm{Q}$ both define the representation used by the auxiliary dynamics objective and provide additional context for action prediction. The dynamics tokens $\bm{Q}$ remain active during inference to preserve the representation pathway learned during training.

\section{Details of Self-Guided World Predictor}
\label{app:sgwp}
\paragraph{Action-Conditioning}
We use an eight-step intervening action sequence to condition latent future prediction, matching the eight-step action chunk generated by the action expert. Each seven-dimensional action is embedded into an action token using an MLP-based projector. A learned temporal position embedding is added to each of the eight action tokens, yielding the action-token sequence $\bm{e}_t^A$. We retain the complete eight-token sequence rather than pooling the action chunk into a single vector. This design preserves the temporal order of the demonstrated transition.

\begin{figure*}[!htbp]
    \centering
    \includegraphics[width=0.6\textwidth]{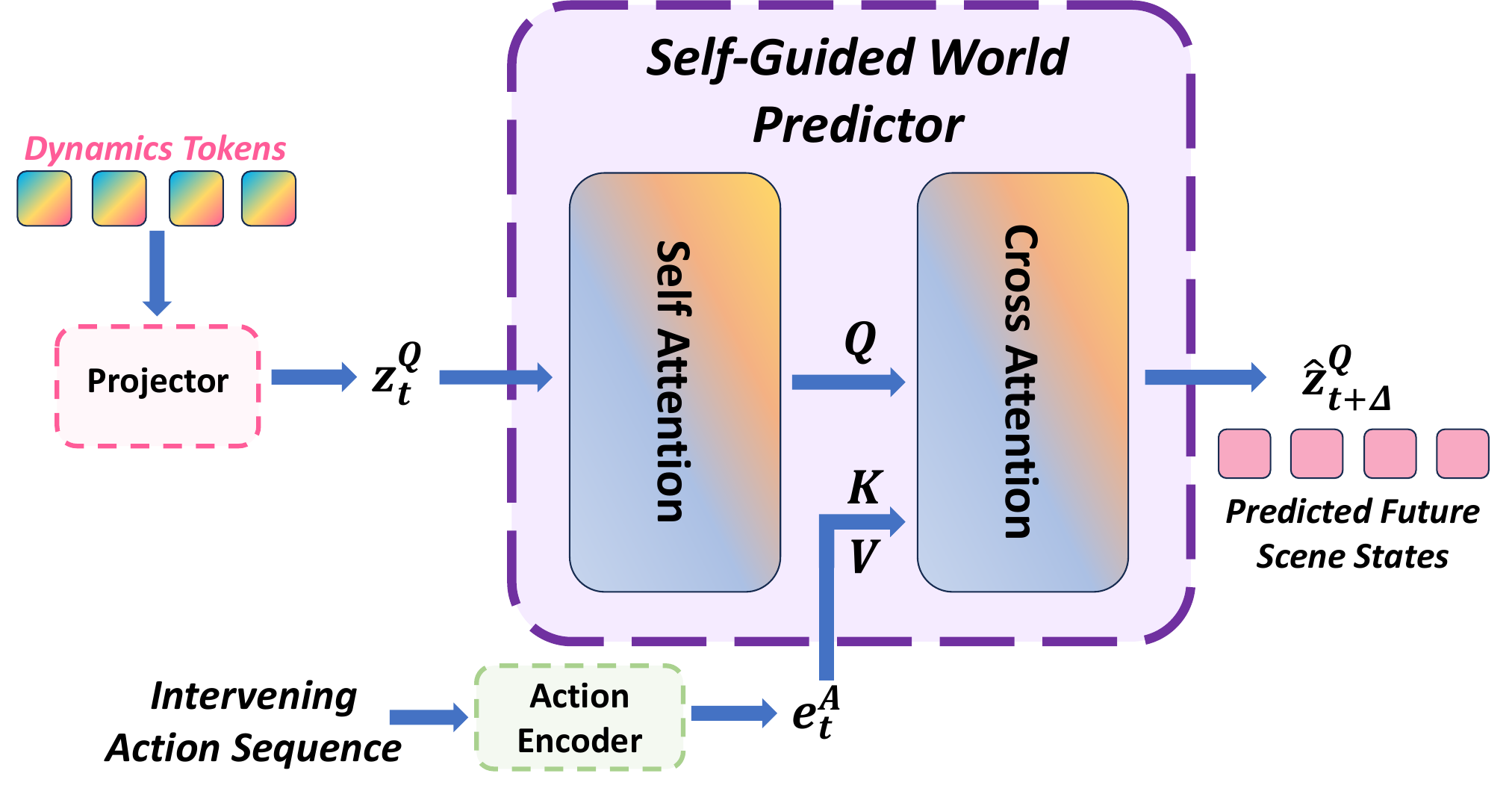}
    \caption{Overview of the Self-Guided World Predictor (SGWP). The projected dynamics-token states are first contextualized through self-attention and then attend to the encoded intervening actions through cross-attention to predict future latent states.}
    \label{fig:SGWP}
\end{figure*}

\paragraph{Future-State Prediction}
The $\bm{H}_t^Q$ is first encoded by the lightweight projector to a compact prediction space ${\bm{z}}_t^{Q} =\mathcal{P}_{\phi}(\bm{H}_t^{Q})$, $\bm{z}_t^{Q} \in \mathbb{R}^{N_q\times d}$, where $\mathcal{P}$ is the projector and $\phi$ is the parameters of this projector. Then $\bm{z}_t^{Q}$ and $\bm{e}_t^A$ are sent to the Self-Guided World Predictor (SGWP) $\mathcal{F}_{\psi}$ to predict future scene state ${\hat{\bm{z}}_{t+\Delta}^{Q}=\mathcal{F}_\psi({\bm{z}}_t^{Q}, \bm{e}_t^A})$, $\hat{\bm{z}}_{t+\Delta}^{Q}\in\mathbb{R}^{N_q\times d}$. The SGWP contains self-attention and cross-attention. As illustrated in Figure~\ref{fig:SGWP}, the current scene states $\bm{z}_t^{Q}$ first interact through self-attention and then act as queries in cross-attention over the action tokens. Self-attention allows these slots to exchange information about the current latent scene, while cross-attention lets each slot selectively condition on the action steps that are most relevant to its predicted change.

\paragraph{EMA target pathway.}
The target future scene states are generated by exponential moving average(EMA) copies of the online VLM backbone $\mathcal{E}_{\theta}$, projector $\mathcal{P}_{\phi}$ and learnable dynamics token $Q$. Following BYOL~\cite{grill2020bootstrap}, their parameters are updated at optimization step $k$ as
\begin{equation}
\bar{\theta}_{k}
=
\mu\bar{\theta}_{k-1}
+
(1-\mu)\theta_{k},
\qquad
\bar{\phi}_{k}
=
\mu\bar{\phi}_{k-1}
+
(1-\mu)\phi_{k},
\qquad
\bar{Q}_{k}
=
\mu\bar{Q}_{k-1}
+
(1-\mu)Q_{k},
\qquad
\mu=0.999.
\end{equation}
Here, $(\theta,\phi,Q)$ and $(\bar{\theta},\bar{\phi},\bar{Q})$ denote
the online and EMA parameters, respectively. At each optimization step, the target parameters retain a fraction $\mu$ of their previous values and incorporate a fraction $1-\mu$ of their online counterparts.

The target pathway is excluded from gradient-based optimization and is updated exclusively through the EMA rule. It is kept in evaluation mode and executed without gradient tracking when computing the target representations. 

Given the future visual observation $\bm{f}_{t+\Delta}^V$and the same language instruction $\bm{f}^L$, and the EMA dynamics token $\bar{\bm{Q}}$, the target future scene states are computed as
\begin{equation}
\bar{\bm{z}}_{t+\Delta}^{Q}=\mathcal{P}_{\bar{\phi}} \left(\bar{\bm{H}}_{t+\Delta}^{Q} \right),
\qquad\bar{\bm{H}}_{t+\Delta}^{Q}=\left[\mathcal{E}_{\bar{\theta}}\left([\bm{f}_{t+\Delta}^{V},\bm{f}^{L},\bar{\bm{Q}}]\right)\right]_{Q},
\end{equation}
where $[\cdot]_{Q}$ selects the last hidden states $\bar{\bm{H}}_{t+\Delta}^{Q}$ associated with the dynamics tokens.

\section{Implementation Details of Geometric Supervision}
\label{app:geo}
We use a frozen VGGT-1B model as the geometric teacher. The teacher is applied only to the main-view observation and is used exclusively for constructing training targets. Specifically, each main-view frame is resized to $518\times518$ using bicubic interpolation with antialiasing and is processed by VGGT as a single-frame sequence. We extract the representation from the final aggregator layer, remove the special tokens, and reshape the remaining $1{,}369$ patch tokens into a $37\times37$ spatial grid. Adaptive average pooling is then applied to reduce the grid to $8\times8$, producing the geometric target $Z_t^G \in \mathbb{R}^{N_v^m\times D_g}$, where $N_v^m$ is the token number of main-view image, $D_g$ is the feature dimension of VGGT.

Geometric supervision is applied directly to the existing main-view image-token states of the policy and does not introduce additional VGGT tokens into the VLM sequence. Since the main-view tokens precede the wrist-view tokens in the policy input sequence, we select the first $N_v^m$ image-token states $\bm{H}_t^{V,m}\in\mathbb{R}^{N_v^m\times D}$, D is the feature dimension of VLM.

Each policy image token is independently mapped to the VGGT
feature dimension using a lightweight token-wise projector
$\mathcal G_{\gamma}$:
\begin{equation}
\hat Z_{t}^{G}
=\mathcal G_{\gamma}\left(\bm{H}_t^{V,m}\right),
\end{equation}

\section{Training Objectives}
We optimize the predictor using the normalized mean-squared error
between the predicted and target future states

\begin{equation} \mathcal{L}_{\mathrm{pred}} = \frac{1}{N_q d} \left\| \hat{\bm{z}}_{t+\Delta}^{Q} - \operatorname{sg} \left( \bar{\bm{z}}_{t+\Delta}^{Q} \right) \right\|_{F}^{2}, 
\end{equation} 
where $N_q$ denotes the number of learnable dynamics tokens, $d$ is the dimensionality of the compact prediction space, and $\operatorname{sg}(\cdot)$ denotes the stop-gradient operation. The normalization by $N_qd$ computes the average prediction error over all dynamics-token slots and feature dimensions. Gradients from $\mathcal{L}_{\mathrm{pred}}$ update the online VLM, learnable dynamics tokens, prediction projector, action encoder, and SGWP, while the EMA target pathway remains gradient-free.

To inject fine-grained spatial structure into the policy representation space, we align the main-view visual-token states with geometric features extracted by a frozen VGGT teacher. Given the main-view observation $\bm{V}_t^{m}$, VGGT produces the geometric features through spatial pooling $\bm{Z}_t^{G}\in\mathbb{R}^{N_v^m\times D_g}$. On the policy side, a lightweight geometry projector maps the corresponding VLM hidden states $\bm{H}_t^{V,m}\in\mathbb{R}^{N_v^m\times D}$to $\hat{\bm{Z}}_t^{G}\in\mathbb{R}^{N_v^m\times D_g}$.
\begin{equation} 
\mathcal{L}_{\mathrm{geo}} = \frac{1}{N_v^m} \sum_{j=1}^{N_v^m} \left[ 1- \cos \left( \hat{\bm{Z}}_{t,j}^{G}, {\bm{Z}}_{t,j}^{G} \right) \right].
\end{equation}
We use directional alignment rather than raw feature regression to avoid over-constraining the magnitude and distribution of the VLM hidden states. This allows the VLM to preserve its original token structure and compatibility with the pretrained representation, while encouraging its image-token states to encode manipulation-relevant 3D cues. By shaping the visual states attended to the dynamics tokens, it encourages the policy-derived predictive space to retain spatial information needed to distinguish action-dependent future transitions.

We train all online components jointly in a single-stage end-to-end optimization procedure. For each training sample, the current observation and language instruction are first processed by the online VLM together with the learnable dynamics tokens. The resulting policy representations are shared by the geometry-alignment branch, the future-state prediction branch, and the conditional flow-matching action expert. The gradients induced by
\begin{equation}
\mathcal{L}
=
\mathcal{L}_{act}
+
\lambda_{geo}\mathcal{L}_{geo}
+
\lambda_{pred}\mathcal{L}_{pred},
\end{equation}
where $\lambda_{\mathrm{geo}}=0.1$ and $\lambda_{\mathrm{pred}}=0.1$ control the contributions of geometry alignment and action-conditioned future-state prediction, respectively. All online components, including the VLM, learnable dynamics tokens, projectors, Self-Guided World Predictor, and action expert, are optimized jointly in a single-stage training procedure. The frozen VGGT teacher receives no gradient updates, while the target VLM and target projector are updated exclusively through EMA. This joint optimization allows spatial supervision, action-conditioned future prediction, and action imitation to shape a common policy representation without requiring separate pretraining or stage-wise fine-tuning.

\section{Simulation Experiments}
\label{app:sim}
The standard LIBERO suites: LIBERO-Spatial, LIBERO-Object, LIBERO-Goal, and LIBERO-Long. These suites evaluate spatial reasoning, object-centric manipulation, goal-conditioned behavior, and long-horizon task execution, respectively. LIBERO-Plus evaluates zero-shot robustness under distribution shifts in camera viewpoint, robot embodiment, language, illumination, background, observation noise, and scene layout. We directly evaluate the policy trained on LIBERO without additional fine-tuning or adaptation.

For SG-WAM, we follow the standard LIBERO and LIBERO-Plus observation, instruction, action-space, and evaluation settings described below. We report model size and embodied-pretraining status.

For LIBERO, we follow the standard protocol and evaluate on the four task suites: LIBERO-Spatial, LIBERO-Object, LIBERO-Goal, and LIBERO-Long. For LIBERO-Plus, we evaluate robustness under distribution shifts. These settings are designed to examine whether the learned policy representations remain stable when visual, semantic, and physical conditions differ from the training distribution.

Our model is initialized from a pretrained vision-language backbone (Qwen3.5-0.8B) and trained end-to-end for visuomotor control. We use 8 dynamics tokens in all main experiments. The proposed self-guided world modeling and geometric supervision objectives are applied only during training. Our model is trained on all four standard LIBERO suites for 40k steps with a global batch size of 96. We adopt a cosine learning-rate schedule with 5k steps linear warmup, with a peak learning rate of 1e-5 for the VLM, 2.5e-5 for the train-only modules, and 1e-4 for action head. No additional fine-tuning is needed on the individual suite.
A rollout is considered successful if the task-specific completion condition is satisfied within the maximum episode horizon.

LIBERO-Plus is designed to test zero-shot robustness by introducing distribution shifts that are not seen during training. Following the benchmark protocol, we evaluate all methods under perturbations in camera viewpoint, robot embodiment, language instruction, illumination, background, observation noise, and scene layout. For all settings, the same checkpoint trained on the original LIBERO simulation data is directly evaluated without any additional fine-tuning or adaptation. The success rate is computed as the percentage of successful rollouts in each perturbation setting, and the final score is obtained by calculating overall successful rollouts. 

Compared with standard LIBERO evaluation, LIBERO-Plus places greater emphasis on whether a policy can preserve task-relevant representations when the observation and instruction distributions change. This setting is particularly challenging for methods whose learned representations are tightly coupled to the training-domain visual appearance or to the surface form of language instructions. Although several baselines perform strongly on the standard LIBERO benchmark, their performance drops noticeably under these zero-shot perturbations, indicating limited robustness beyond the in-distribution evaluation setting.

\section{Real-World Experiments}
\label{app:real}
The real-world evaluation is designed to examine whether the proposed self-guided world modeling framework can transfer from simulation-style policy learning to physical robot manipulation. Unlike simulation benchmarks, real-world execution introduces additional challenges such as sensor noise, imperfect calibration, object pose uncertainty, and contact dynamics.

\begin{figure*}[!htbp]
    \centering
    \includegraphics[width=0.98\textwidth]{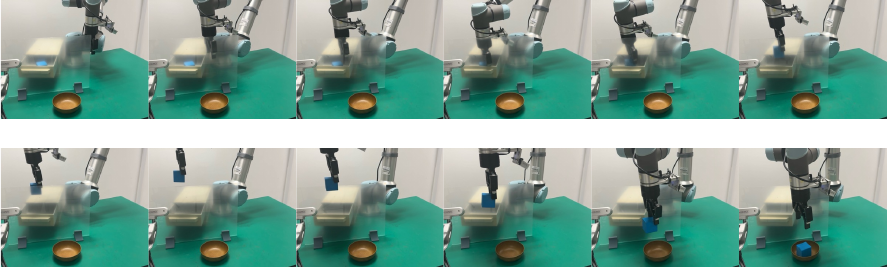}
    \caption{Complete visualization of the Pick and Place task.}
    \label{fig:complete_pick}
\end{figure*}

\begin{figure*}[!t]
    \centering
    \includegraphics[width=0.98\textwidth]{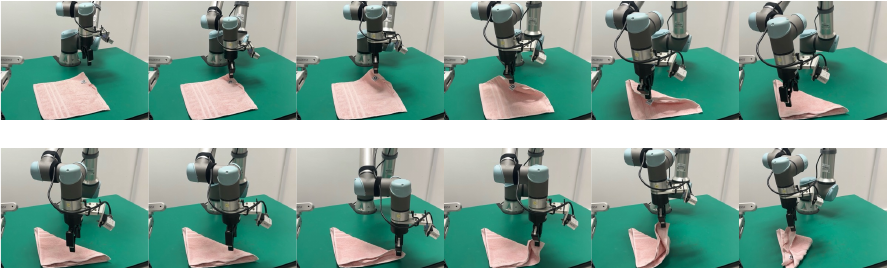}
    \caption{Complete visualization of the Towel Folding task.}
    \label{fig:complete_fold}
\end{figure*}

The in-distribution (ID) setting consists of three tabletop manipulation tasks, with 100 expert demonstrations collected for each task:
\begin{itemize}
    \item \textbf{Pick and Place.} As shown in Figure~\ref{fig:complete_pick}, the robot retrieves a blue cube from an open drawer and places it into a bowl while a barrier obstructs the direct transfer path. Successful execution requires a collision-free trajectory that clears the barrier while maintaining a stable grasp. This task primarily evaluates 3D spatial reasoning and obstacle-aware trajectory planning.
    \item \textbf{Towel Folding.} As shown in Figure~\ref{fig:complete_fold}, the robot folds a flat towel in half twice to produce a compact configuration. 
    Each interaction induces non-rigid deformation and changes the feasible grasping regions for the subsequent fold. Successful execution requires tracking the evolving cloth geometry, selecting precise grasp points, and aligning overlapping layers. This task primarily evaluates deformable-object state prediction and geometry-aware manipulation.
    \item \textbf{Toolbox Organization.} As shown in Figure~\ref{fig:complete_pick}, the robot sequentially places a screwdriver and two gears into a toolbox before closing it. The objects are initially stacked, creating occlusions and dependencies between successive actions. Successful execution requires selecting an appropriate manipulation order, tracking task progress, and maintaining spatial accuracy over an extended sequence. This task primarily evaluates long-horizon planning and multi-object reasoning.
\end{itemize}
\begin{figure*}[!t]
    \centering
    \includegraphics[width=0.98\textwidth]{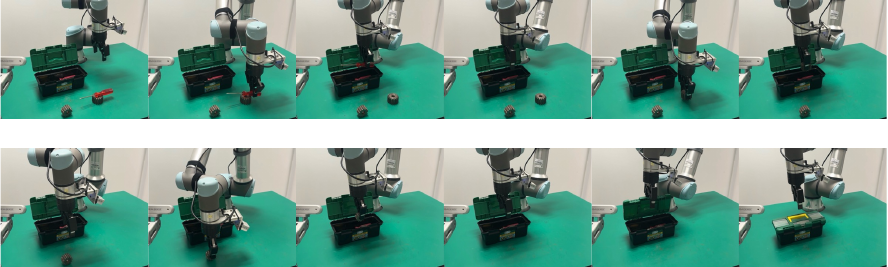}
    \caption{Complete visualization of the Toolbox Organization task.}
    \label{fig:complete_toolbox}
\end{figure*}
These tasks are designed to cover complementary manipulation challenges, including obstacle-aware pick-and-place, deformable-object manipulation, and long-horizon multi-object organization. All demonstrations contain synchronized RGB observations, language instructions, and robot actions. During evaluation, the policy is tested on the same task families.

\begin{figure*}[!htbp]
    \centering
    \includegraphics[width=0.98\textwidth]{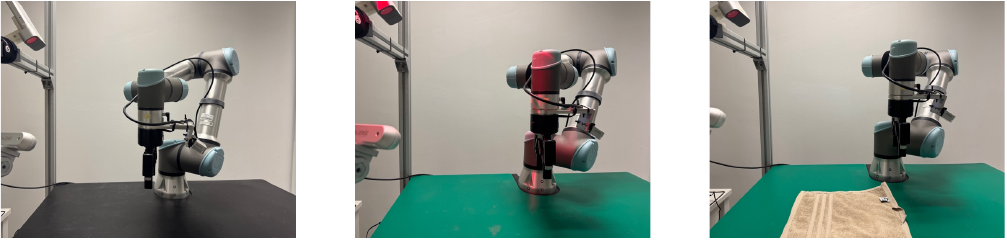}
    \caption{Visualization of 3 OOD settings. 1) \textbf{Left:} Background Shift. 2) \textbf{Middle:} Light Change. 3) \textbf{Right:} Novel Object.}
    \label{fig:ood_setting}
\end{figure*}

As illustrated in Figure~\ref{fig:ood_setting}, the out-of-distribution (OOD) setting consists of three test-time variations that are not represented in the expert demonstrations:
\begin{itemize}
    \item \textbf{Background Shift.} Expert demonstrations are collected with one table covering, which is replaced by a visually distinct covering during evaluation.
    \item \textbf{Light Change.} During evaluation, a programmable light source illuminates the workspace from a fixed position and intensity not encountered during training.
    \item \textbf{Novel Object.} Towel Folding uses an unseen towel, whereas Pick and Place uses cubes with novel colors and correspondingly updated language instructions.
\end{itemize}

The OOD setting evaluation uses the same task definitions and success criteria as the in-distribution setting but introduces test-time variations. This protocol allows us to examine whether the learned policy can preserve task-relevant representations when visual appearance, illumination, or object identity changes. No additional demonstrations, fine-tuning, or test-time adaptation are used for the OOD evaluation. 
All models are trained jointly on all three collected tasks for 40 epochs with a global batch size of 96. The learning rates is the same as simulation experiments. No additional fine-tuning is performed for any individual task or evaluation condition. The same final checkpoint is used for both ID and OOD real-world evaluations, without additional demonstrations, test-time adaptation, or task-specific calibration.

In the real-world experiments, we use a UR5e robotic arm as our main manipulation platform, an Azure Kinect camera for main RGB image acquisition and a RealSense D405 for robot gripper RGB image acquisition. This platform provides a large workspace, high repeatability, and easy programmability. During operation, the robotic arm and the cameras work together through real-time data exchange to achieve precise  manipulation. All devices are connected to a workstation with an NVIDIA RTX 3090 GPU for model inference and control.

Across both settings, SG-WAM achieves the highest average success rate, suggesting that the proposed training objectives improve not only standard imitation performance but also real-world generalization. In the ID setting, the gains of SG-WAM are consistent across all three tasks. This indicates that the improvement is not specific to a particular manipulation type. The consistent improvements across these tasks suggest that SG-WAM provides a general benefit to spatially grounded and temporally coherent policy learning.
The stronger OOD performance of SG-WAM suggests that the learned representations are less tied to superficial visual cues in the training demonstrations. The policy appears to better preserve task-relevant object relations and state transitions under visual distribution shifts. The advantage of SG-WAM is particularly meaningful for Toolbox Organization. Unlike single-step manipulation, this task requires multiple successful subtasks to be completed in sequence. Errors in early stages, such as inaccurate grasping or object placement, can accumulate and prevent later stages from succeeding. The improved performance on this task therefore suggests that action-conditioned latent prediction helps maintain task-progress information over extended manipulation horizons. Together, these results provide additional evidence that self-guided world modeling improves both real-world execution stability and robustness to unseen evaluation conditions.

\begin{table*}[!htbp]
\centering
\small
\setlength{\tabcolsep}{2.5pt}

\begin{tabular}{@{}l cc cccc@{}}
\toprule
\multicolumn{1}{c}{\multirow[c]{2}{*}{\textbf{Model}}}
& \multicolumn{2}{c}{\textbf{Towel Folding}}
& \multicolumn{4}{c}{\textbf{Toolbox Organization}} \\
\cmidrule(lr){2-3}
\cmidrule(lr){4-7}
& First Folding & Second Folding & Pick the Screwdriver & Pick the First Gear & Pick the Second Gear & Close the Toolbox \\
\midrule
VLA-JEPA & 50\% & 20\% & 50\% & 40\% & 20\% & 20\%\\
VPP & 60\% & 35\% & 70\% & 50\% & 40\% & 30\% \\
\midrule
\textbf{\model} & \textbf{75\%} & \textbf{45\%} & \textbf{80\%} & \textbf{60\%} & \textbf{50\%} & \textbf{50\%}\\
\bottomrule
\end{tabular}
\caption{Subtasks success rates of Towel Folding and Toolbox Organization tasks in ID settings. \textbf{Bold} indicates the best result.}
\label{tab:realworld_subtask}
\end{table*}

Table~\ref{tab:realworld_subtask} provides a fine-grained breakdown of the Towel Folding and Toolbox Organization tasks. Across all subtasks, SG-WAM consistently outperforms VLA-JEPA and VPP, suggesting that the gains are not only reflected in final task completion but also in intermediate execution stages. 

For the Towel Folding, all methods achieve higher success rates on the first fold than on the second fold, indicating that the second fold is more challenging because the policy must reason over the deformed cloth state induced by the first interaction. Compared to other baselines, SG-WAM shows better robustness in sequential deformable-object manipulation.

For Toolbox Organization, the success rates decrease as the task progresses from picking the screwdriver to picking the gears and closing the toolbox. This trend reflects the long-horizon nature of the task, where early grasping or placement errors can accumulate and affect later subtasks. SG-WAM achieves larger improvements in the later stages. These results indicate that the proposed self-guided world modeling objective helps maintain more temporally coherent task representations during multi-step real-world manipulation.

\section{Ablation Study}
\label{app:ablation}
\begin{figure*}[!htbp]
    \centering
    \includegraphics[width=0.98\textwidth]{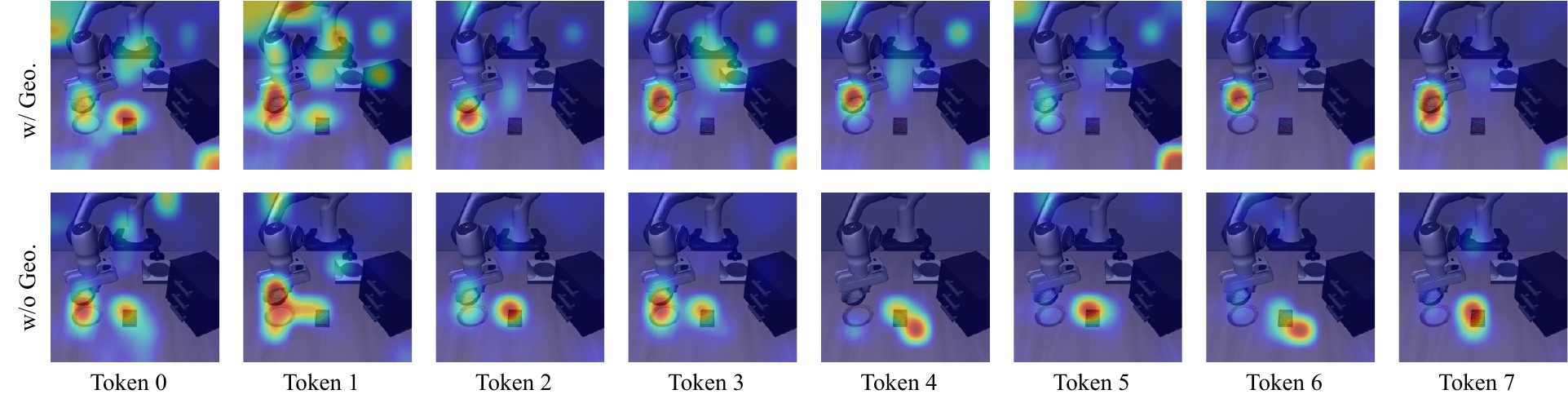}
    \caption{Visualization of the attention weight matrix of all latent dynamics tokens attending to main-view image tokens. w/ Geo. and w/o Geo. denote with geometric supervision and without geometric supervision, respectively.}
    \label{fig:complete_attention_map}
\end{figure*}

\paragraph{Complete Attention Maps Across Dynamics Tokens}
Figure~\ref{fig:complete_attention_map} provides the complete attention visualization for all dynamics tokens. While the main paper reports representative tokens for readability, here we include the full set of eight dynamics tokens to examine whether the observed pattern is consistent across tokens. With geometric supervision, the dynamics tokens attend more consistently to interaction-relevant regions, including the robot end effector, the manipulated object, the target receptacle, and the surrounding workspace. In contrast, without geometric supervision, several tokens place stronger attention on isolated visually salient regions that are less directly related to the robot-object interaction. These observations are consistent with the analysis in the main paper and further suggest that geometric supervision shapes policy visual representations into a more manipulation-relevant spatial context, thereby providing a structured basis for action-conditioned latent dynamics modeling.

\paragraph{Effect of Intervening Action Conditioning}
\begin{table}[!htbp]
\centering
\small
\begin{tabular}{lccccc}
\toprule
Variant & Spatial & Object & Goal & Long & Avg. \\
\midrule
Null-Action Sequence
& 98.4 & 99.2 & 98.0 & 94.6 & 97.6 \\
SG-WAM
& \textbf{99.4} & \textbf{99.8} & \textbf{98.6}
& \textbf{96.2} & \textbf{98.5} \\
\bottomrule
\end{tabular}
\caption{Ablation of the action information provided to the Self-Guided World Predictor on LIBERO. For the null-action variant, the ground-truth intervening action sequence is replaced with an all-zero sequence before being processed by the action encoder. The action-conditioning architecture, temporal positional embeddings, and all other model components and training settings remain unchanged. Success rates (\%) are reported.}
\label{tab:action_conditioning}
\end{table}

To investigate whether the actual intervening actions provide useful information for predicting future policy states, we replace he ground-truth action sequence $\mathbf{A}^{(\Delta)}_t$ with an all-zero sequence of the same shape:
\begin{equation}
\widetilde{\mathbf{A}}^{(\Delta)}_t
=
\mathbf{0}.
\end{equation}

The zero-valued sequence is subsequently processed by the same
action encoder used in the full model:

\begin{equation}
\widetilde{\mathbf{e}}^A_t
=
\mathcal{A}_{\eta}
\left(
\widetilde{\mathbf{A}}^{(\Delta)}_t
\right),
\qquad
\hat{\mathbf{z}}^Q_{t+\Delta}
=
F_{\psi}
\left(
\mathbf{z}^Q_t,
\widetilde{\mathbf{e}}^A_t
\right).
\end{equation}
Thus, this variant preserves the action-conditioning pathway, sequence length, and temporal positional embeddings, while removing information about the actual robot actions executed between the current and future observations. All other model components, objectives, and training settings remain unchanged.

As shown in Table~\ref{tab:action_conditioning}, replacing the intervening actions with a null-action sequence reduces the average success rate from 98.5\% to 97.6\%. Performance decreases on all four LIBERO suites, with reductions of 1.0, 0.6, 0.6, and 1.6 percentage points on LIBERO-Spatial, LIBERO-Object, LIBERO-Goal, and LIBERO-Long, respectively. This result indicates that the specific intervening actions provide useful information for predicting how the current policy state evolves, beyond the temporal structure and architectural capacity retained by the null-action variant.

The null-action variant remains competitive, indicating that the current policy state already contains substantial predictive information. Nevertheless, the consistent gains from using the ground-truth intervening actions show that specific action information improves the modeling of future latent transitions, particularly for long-horizon tasks.

\newpage
\section{Real-World OOD Rollouts Visualization}
\begin{figure*}[!htbp]
    \centering
    \includegraphics[width=0.98\textwidth]{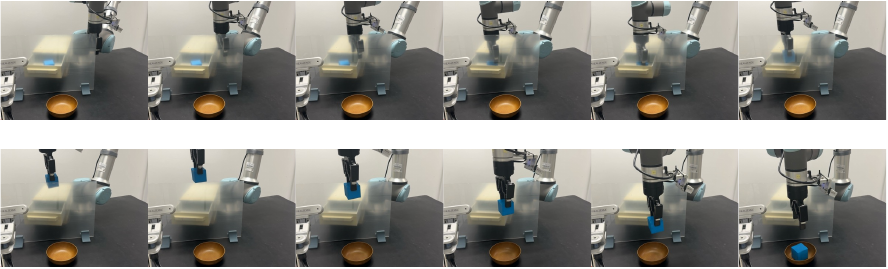}
    \caption{Complete visualization of the Pick and Place task under the background shift.}
    \label{fig:pick_background}
\end{figure*}

\begin{figure*}[!htbp]
    \centering
    \includegraphics[width=0.98\textwidth]{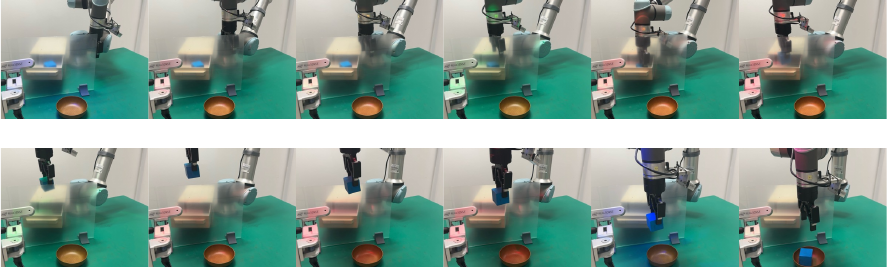}
    \caption{Complete visualization of the Pick and Place task under the light change.}
    \label{fig:pick_light}
\end{figure*}

\begin{figure*}[!htbp]
    \centering
    \includegraphics[width=0.98\textwidth]{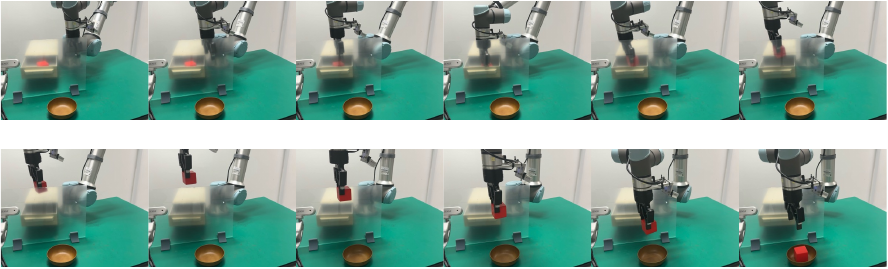}
    \caption{Complete visualization of the Pick and Place task under the novel object.}
    \label{fig:pick_novel_object}
\end{figure*}

\begin{figure*}[!htbp]
    \centering
    \includegraphics[width=0.98\textwidth]{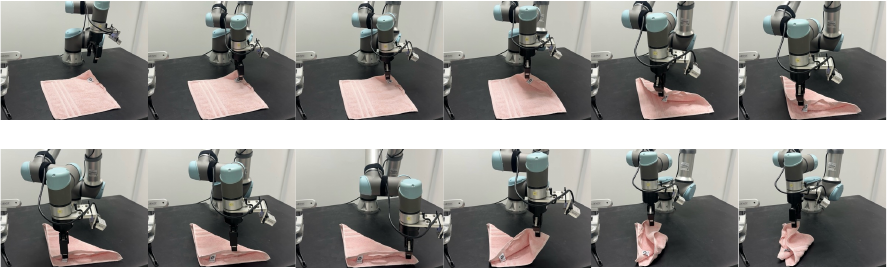}
    \caption{Complete visualization of the Towel Folding task under background shift.}
    \label{fig:fold_background}
\end{figure*}

\begin{figure*}[!htbp]
    \centering
    \includegraphics[width=0.98\textwidth]{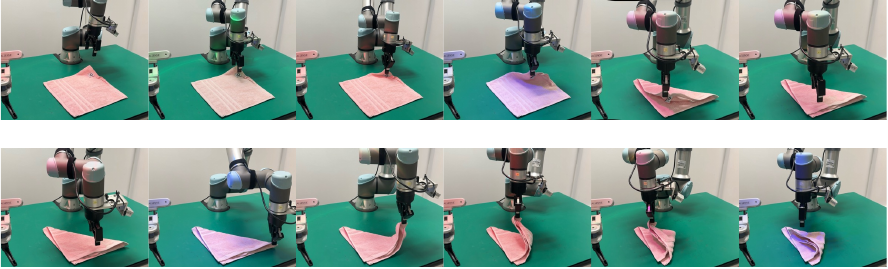}
    \caption{Complete visualization of the Towel Folding task under light change.}
    \label{fig:fold_light}
\end{figure*}

\begin{figure*}[!htbp]
    \centering
    \includegraphics[width=0.98\textwidth]{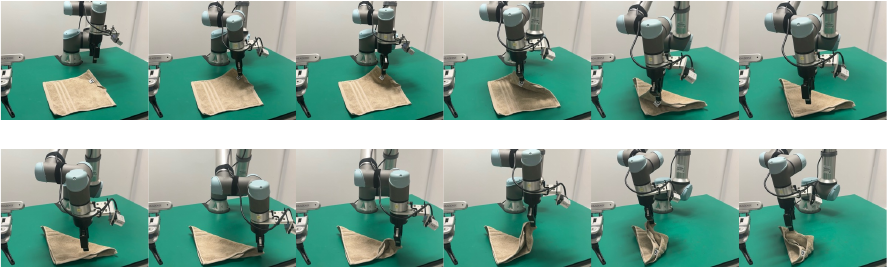}
    \caption{Complete visualization of the Towel Folding task under novel object.}
    \label{fig:fold_novel_object}
\end{figure*}
\newpage
\section{LIBERO Visualization}

\begin{figure*}[!htbp]
    \centering
    \includegraphics[width=0.98\textwidth]{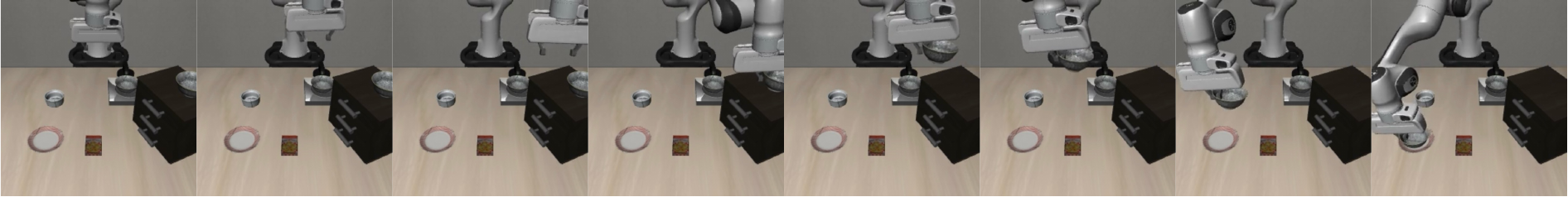}
    \caption{Visualization of the LIBERO-Spatial suite: Pick up the black bowl on the wooden cabinet and place it on the plate.}
    \label{fig:spatial}
\end{figure*}
\begin{figure*}[!htbp]
    \centering
    \includegraphics[width=0.98\textwidth]{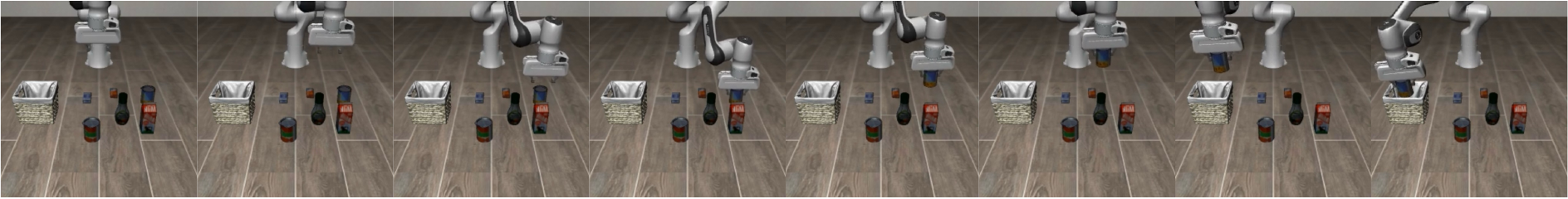}
    \caption{Visualization of the LIBERO-Object suite: Pick up the alphabet soup and place it in the basket.}
    \label{fig:object}
\end{figure*}
\begin{figure*}[!htbp]
    \centering
    \includegraphics[width=0.98\textwidth]{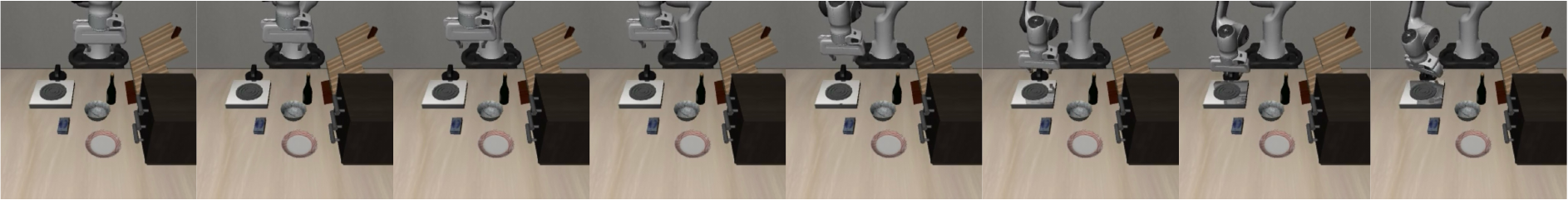}
    \caption{Visualization of the LIBERO-Goal suite: Turn on the stove.}
    \label{fig:goal}
\end{figure*}
\begin{figure*}[!htbp]
    \centering
    \includegraphics[width=0.98\textwidth]{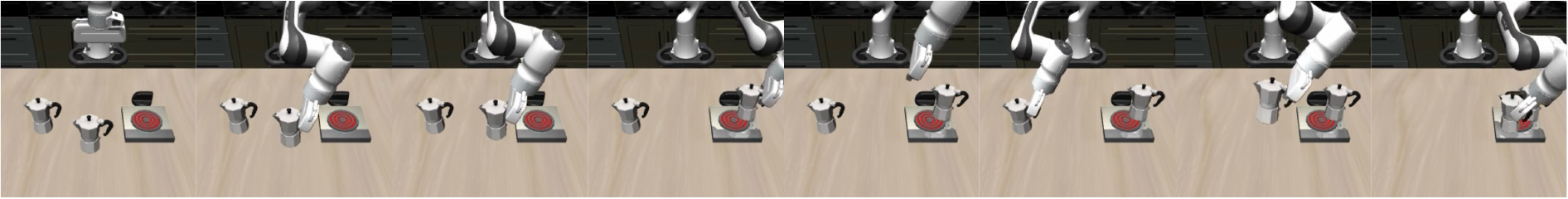}
    \caption{Visualization of the LIBERO-Long suite: Put both moka pots on the stove.}
    \label{fig:long}
\end{figure*}
\section{LIBERO-Plus Visualization}

\begin{figure*}[!htbp]
    \centering
    \includegraphics[width=0.98\textwidth]{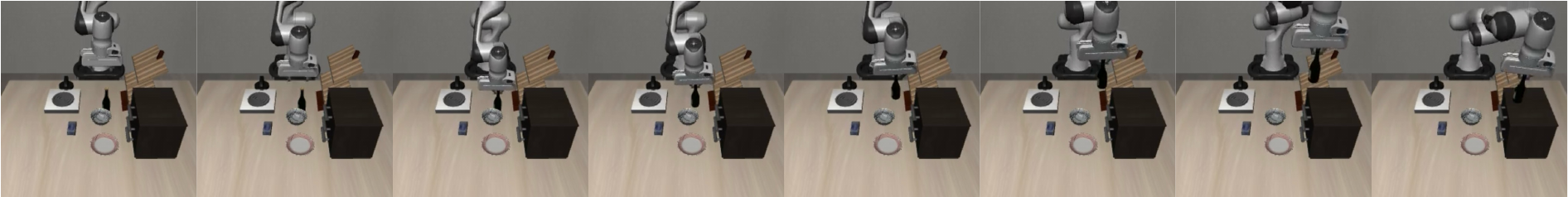}
    \caption{Visualization of the perturbation of  Camera Viewpoints: Put the wine bottle on top of the cabinet.}
    \label{fig:camear_plus}
\end{figure*}

\begin{figure*}[!htbp]
    \centering
    \includegraphics[width=0.98\textwidth]{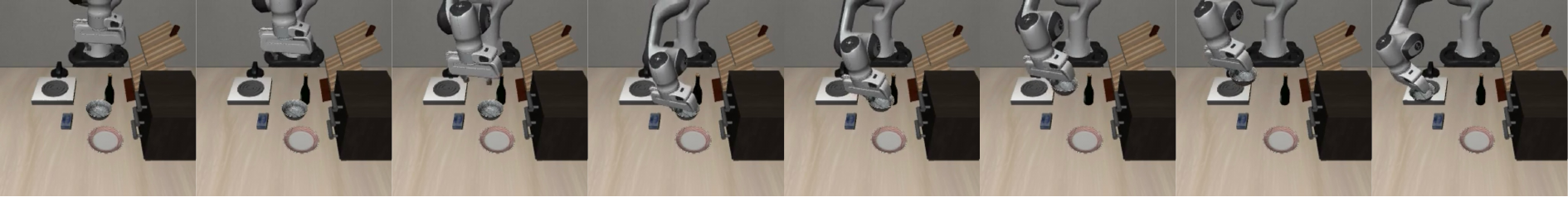}
    \caption{Visualization of the perturbation of Robot Initial States: Put the bowl on the stove.}
    \label{fig:robot_plus}
\end{figure*}

\begin{figure*}[!htbp]
    \centering
    \includegraphics[width=0.98\textwidth]{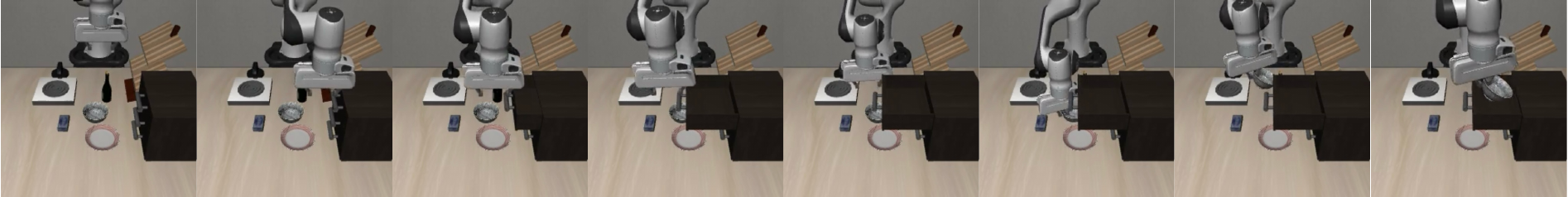}
    \caption{Visualization of the perturbation of  Language Instructions: Your next task is simple: open that top drawer and place the bowl inside neatly.}
    \label{fig:language_plus}
\end{figure*}

\begin{figure*}[!htbp]
    \centering
    \includegraphics[width=0.98\textwidth]{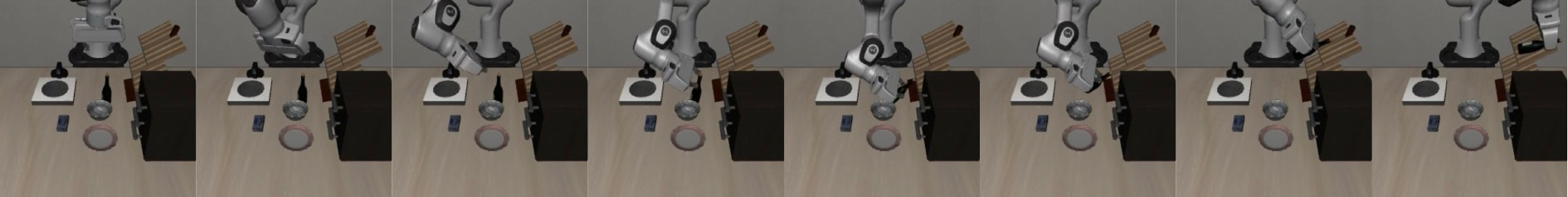}
    \caption{Visualization of the perturbation of Light Conditions: Put the wine bottle on the rack.}
    \label{fig:light_plus}
\end{figure*}

\begin{figure*}[!htbp]
    \centering
    \includegraphics[width=0.98\textwidth]{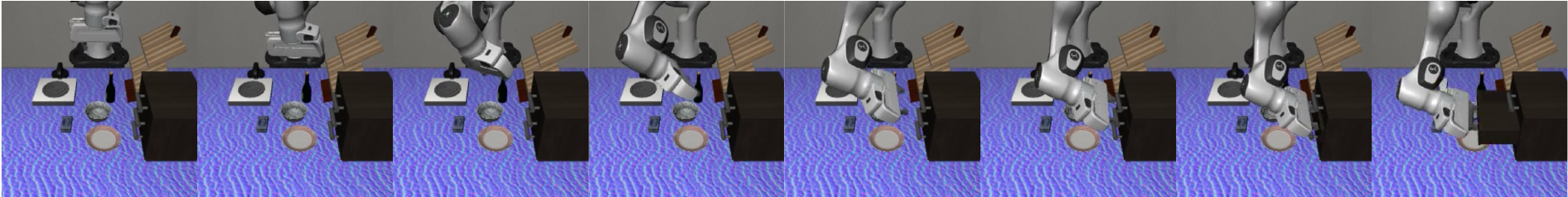}
    \caption{Visualization of the perturbation of Background Textures: Open the middle drawer of the cabinet.}
    \label{fig:background_plus}
\end{figure*}

\begin{figure*}[!htbp]
    \centering
    \includegraphics[width=0.98\textwidth]{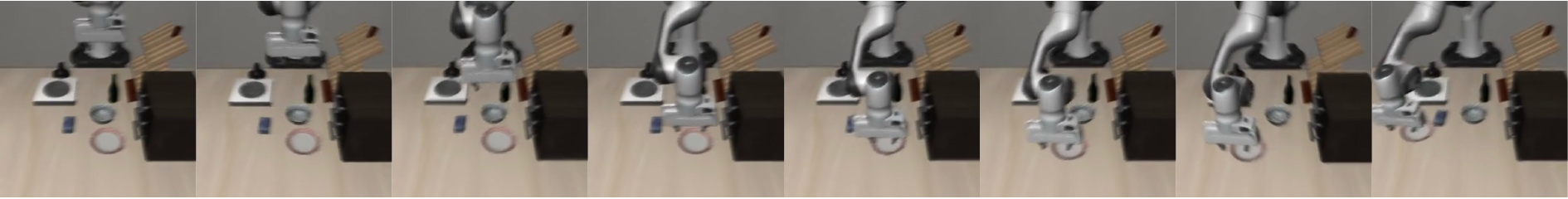}
    \caption{Visualization of the perturbation of Sensor Noise: Push the plate to the front of the stove.}
    \label{fig:noise_plus}
\end{figure*}

\begin{figure*}[!htbp]
    \centering
    \includegraphics[width=0.98\textwidth]{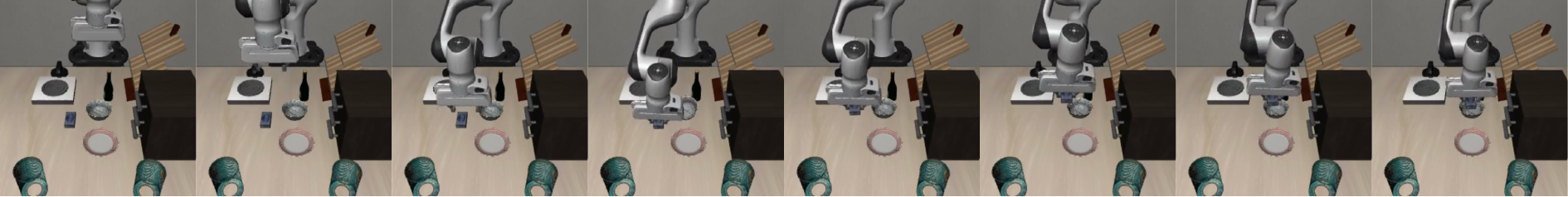}
    \caption{Visualization of the perturbation of Objects Layout: Put the cream cheese in the bowl.}
    \label{fig:layout_plus}
\end{figure*}

\end{document}